%% file: main.tex
\documentclass{article} % For LaTeX2e
\usepackage{iclr2027_conference,times}

\input{math_commands.tex}

\usepackage{hyperref}
\usepackage{url}
\usepackage{diagbox}
\usepackage{multirow}
\usepackage{pifont}
\usepackage[table]{xcolor}
\usepackage{amsmath}
\usepackage{amssymb}
\usepackage{graphicx}
\usepackage{booktabs}
\usepackage{subcaption, caption}
\usepackage{wrapfig}
\usepackage{cleveref} 
\crefname{figure}{Figure}{Figs.}
\crefname{table}{Table}{Tabs.}
\crefname{section}{Section}{Secs.}

\title{DeCo: Efficient Decouple-to-Couple Learning for Multi-Task Visual Grounding}

\author{Xiaoqiang Lu, Licheng Jiao\thanks{Corresponding author.}, Long Sun, Yuting Yang, Xu Liu, Lingling Li, Wenping Ma, Fang Liu \\
Key Laboratory of Intelligent Perception and Image Understanding of Ministry of Education\\
School of Artificial Intelligence, Xidian University \\
Xi'an, Shaanxi 710071, China \\
\texttt{xqlu@xidian.edu.cn, lchjiao@mail.xidian.edu.cn} \\
}

\iclrfinalcopy % Uncomment for camera-ready version, but NOT for submission.
\begin{document}

\maketitle

\input{Sections/0_abstract}

\input{Sections/1_introduction}

\input{Sections/2_related_work}

\input{Sections/3_methodology}

\input{Sections/4_experiments}

\input{Sections/5_conclusion}

\newpage

\subsection*{AI use statement}

In this work, we used generative AI tools for language polishing, grammar error checking, and supporting qualitative and quantitative data analysis. We have not used generative AI tools for generating synthetic data sets, helping develop theoretical models, and other required disclosure tasks are not applicable to this work.

\subsection*{Ethics statement}

This submission adheres to the ICLR Code of Ethics and Code of Conduct. Our work does not involve human subjects or animal experiments; thus, no IRB approval is required.
All datasets used are publicly available for research under their respective licenses, which we follow without redistributing restricted content.

\subsection*{Reproducibility statement}

To ensure full reproducibility of our experimental results, we provide detailed algorithmic and architectural specifications in both the main paper and the supplementary material. After the review process, we will release the complete source code, training and testing scripts, and all trained model checkpoints. We believe that this combination of paper and code supports end-to-end reproducibility and facilitates fair comparison with future work.

% \subsubsection*{Author Contributions}
% If you'd like to, you may include  a section for author contributions as is done
% in many journals. This is optional and at the discretion of the authors.

\subsubsection*{Acknowledgments}

This work was supported by the Postdoctoral Fellowship Program of China Postdoctoral Science Foundation under Grant Number GZC20261905, the China Postdoctoral Science Foundation under Grant Number 2026M791683 and 2025M781511, and the Natural Science Basic Research Program of Shaanxi, China (2026JC-YBQN-0813).

\bibliography{iclr2027_conference}
\bibliographystyle{iclr2027_conference}

\newpage
\appendix
\input{Sections/6_appendix}

\end{document}

%% file: math_commands.tex
\usepackage{amsmath,amsfonts,bm}

\def\eqref#1{equation~\ref{#1}}
\def\1{\bm{1}}

\DeclareMathAlphabet{\mathsfit}{\encodingdefault}{\sfdefault}{m}{sl}
\SetMathAlphabet{\mathsfit}{bold}{\encodingdefault}{\sfdefault}{bx}{n}

%% file: Sections/0_abstract.tex
\begin{abstract}

Multi-task visual grounding requires models to jointly understand linguistic semantics and perform accurate visual localization and segmentation. Despite the success of multimodal large language models, effectively adapting them to multiple grounding objectives remains challenging. Existing methods commonly enforce task cooperation through shared representations, while overlooking the intrinsic conflict between task-oriented feature interests. In this paper, we introduce \textbf{DeCo}, an efficient \textbf{De}couple-to-\textbf{Co}uple learning framework that resolves this dilemma through a two-stage paradigm: task-specific representation decoupling followed by complementary prior coupling. Specifically, we first propose Task-aware Semantic Decoupling (TSD) to route shared visual cues into individual features under salient word-level guidance, alleviating representation interference between localization and segmentation. Furthermore, we observe that segmentation naturally provides informative localization priors due to dense supervision. Based on this insight, we introduce Hybrid Prior Coupling (HPC), which integrates sentence-level semantic prior with mask-derived spatial prior for enhanced grounding. Built upon a frozen multimodal encoder, DeCo requires lightweight trainable parameters while achieving strong generalization across multiple grounding objectives. Extensive experiments on RefCOCO/+, G-Ref, ReferIt, Flickr, DIOR-RSVG, SARVG1.0, RRSIS-D, RIS-LAD, and RefDIOR demonstrate that DeCo achieves state-of-the-art performance on both natural and remote sensing benchmarks. The code and models are available at \url{https://github.com/xiaoqiang-lu/DeCo}.

\end{abstract}

%% file: Sections/1_introduction.tex
\section{Introduction}

Multi-task visual grounding (MTVG) aims to simultaneously comprehend a natural language description and identify its corresponding visual target through multiple grounding objectives~\citep{polyformer, oneref, latent, instancevg}, primarily including referring expression comprehension (REC)~\citep{transvg++, hivg, scanformer, simvg} and referring expression segmentation (RES)~\citep{lavt, remamber, safire, amlris, wimfris}. By bridging linguistic concepts and visual content, MTVG serves as a fundamental capability for vision-language understanding~\citep{tpami_survey, res_survey}. Recent advances in multimodal large language models (MLLMs)~\citep{llava, internvl2.5, deepseekvl2} have substantially improved grounding performance by leveraging informative, aligned cross-modal representations learned from extensive pre-training~\citep{gsva, lion, segllm}. However, effectively adapting these foundation models to multiple grounding objectives remains challenging, especially when different tasks require distinct and sometimes conflicting visual representations~\citep{rrsecs, grex}.

\begin{figure*}[t]
    \centering
    \vspace{-5pt}
    \includegraphics[width=1.0\linewidth]{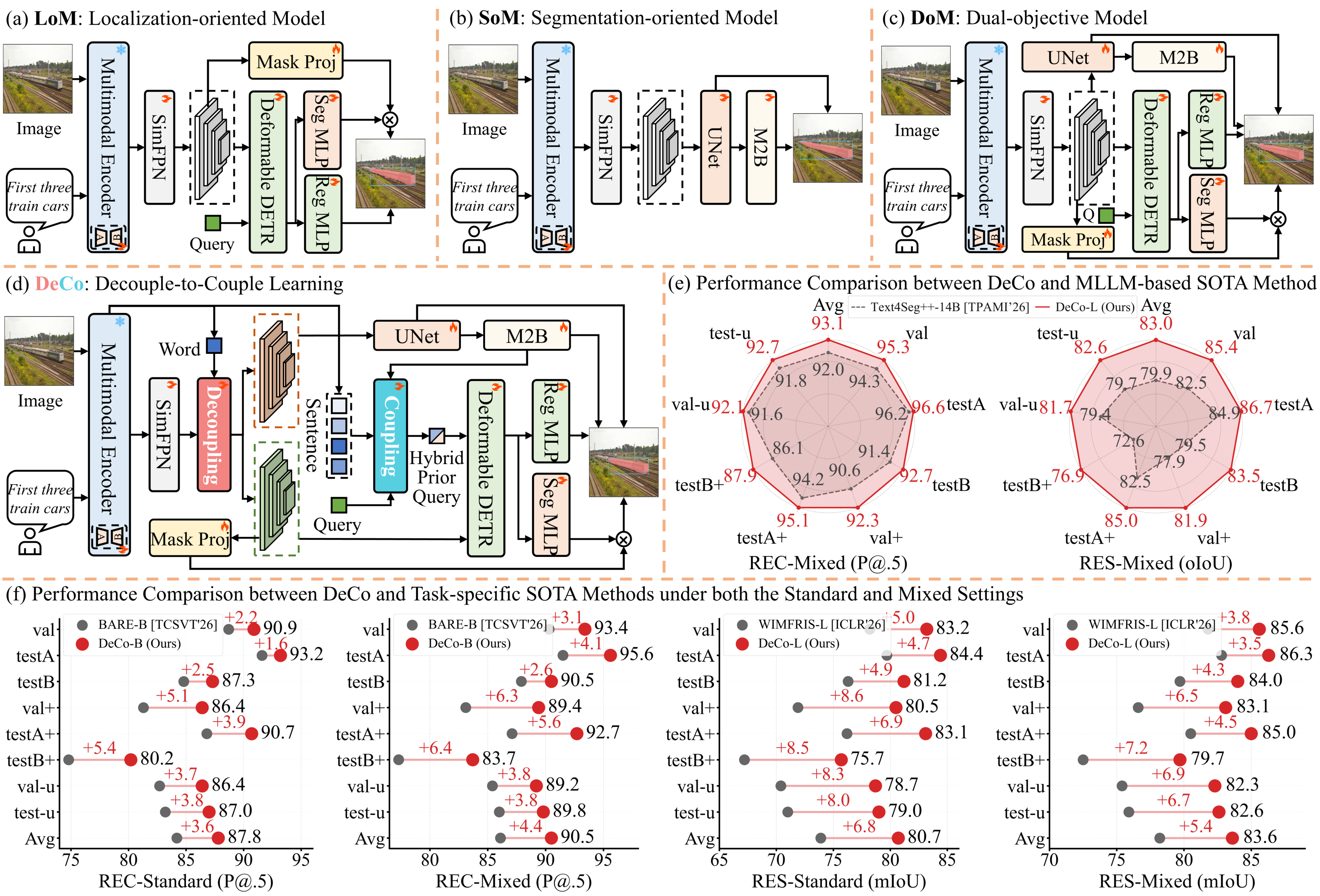}
    \caption{Overview. (a) LoM, which adopts a DETR-based detector with a query-based mask projection branch. (b) SoM, which employs a UNet decoder followed by a mask-to-box (M2B) operation. (c) DoM, which integrates LoM and SoM upon the shared multi-scale features. (d) DeCo, which follows a decouple-to-couple learning paradigm to disentangle discriminative representations and harness hybrid priors for MTVG. (e) Comparison between DeCo and MLLM-based method, Text4Seg++~\citep{text4seg++}, on both REC and RES. (f) Comparison between DeCo and REC-specific method, BARE~\citep{bare}, and RES-specific method, WIMFRIS~\citep{wimfris}.}
    \label{fig_overview}
    \vspace{-10pt}
\end{figure*}
    
Existing methods for MTVG can be broadly categorized into two paradigms. The first one leverages MLLM as a multimodal encoder, which typically regresses the coordinates of the referred object~\citep{vpp, sifthinker, vgent} for REC, and employs various strategies, such as prompting a special token into the mask decoder~\citep{lisa, unipixel, fclm} of SAM~\citep{sam, sam2}, regressing sampled contour points~\citep{visionllm, vistallm, visionllmv2}, or generating a text label corresponding to its image patch~\citep{text4seg, text4seg++}, for RES. These methods exploit the strong reasoning capability of MLLM but come with billions of parameters and additional post-processing procedures. The other one employs pretrained vision-language foundation models as multimodal encoders~\citep{clip, beit3} and designs task-specific architectures to perform REC and RES~\citep{eevg, propvg, c3vg}. This paradigm capitalizes on task-oriented preference modeling but typically requires fully training the entire model. However, both of them fundamentally rely on shared visual representations across distinct tasks. Despite this design facilitating knowledge transfer and reducing model complexity, it implicitly assumes that different objectives can benefit from a unified latent space. In practice, localization emphasizes high-level semantic correspondence and object-level spatial reasoning, whereas segmentation requires fine-grained spatial details and pixel-level discrimination~\citep{rrsecs, grex}.

\label{sec_introduction}
To examine this assumption, we first construct three simple yet effective baselines based on a frozen multimodal encoder~\citep{beit3} with low-rank adaptation~\citep{lora}, followed by a feature pyramid neck~\citep{vitdet} that expands a single-layer encoder output into multi-scale features for MTVG, as illustrated in \cref{fig_overview}: (a) The Localization-oriented Model (LoM) interacts a randomly initialized object query with multi-scale features through a deformable decoder~\citep{deformabledetr} to perform object-level localization, while projecting the output query and multiplying it with the high-resolution feature to generate its corresponding pixel-level segmentation; (b) The Segmentation-oriented Model (SoM) feeds multi-scale features into a UNet decoder~\citep{unet} to predict a global mask, from which the minimal enclosing bounding box is subsequently derived; (c) By integrating LoM and SoM built upon shared multi-scale features, the Dual-objective Model (DoM) jointly optimizes box regression and mask prediction within a unified space. Although the results in \cref{tab_abla_comp} demonstrate that all three baselines already outperform recent methods BARE~\citep{bare} and AMLRIS~\citep{amlris} on REC and RES, respectively, we observe that LoM achieves stronger performance on REC but is relatively weaker on RES, whereas SoM exhibits the opposite trend, indicating that REC and RES have different feature preferences. More importantly, DoM largely inherits the respective strengths of both, yet fails to obtain substantial additional gains from multi-task learning. This finding suggests that simply combining multiple task heads is insufficient to fully exploit task complementarity. Instead, the shared representation itself may hinder each objective from developing its own task-oriented feature interests.

In this paper, we propose \textbf{DeCo}, an efficient \textbf{De}couple-to-\textbf{Co}uple learning framework to address the above limitation, as shown in \cref{fig_overview} (d). Unlike existing methods~\citep{polyformer, eevg, oneref, c3vg, latent, instancevg} which directly optimize multiple objectives over shared visual space, DeCo introduces a two-stage paradigm that separates task-specific representation learning from cross-objective prior exploitation, allowing each task to preserve its own feature preferences while benefiting from complementary knowledge. First, we propose \textbf{Task-aware Semantic Decoupling (TSD)} to transform shared multi-scale features into individual representations for localization and segmentation, respectively. TSD leverages a salient word-level feature to modulate visual features and employs lightweight experts to extract task-oriented feature interests of REC and RES. This design enables each task to learn characteristics better aligned with its preferences while retaining the efficiency of shared multi-scale encoding. Second, we introduce \textbf{Hybrid Prior Coupling (HPC)} to enrich the object query with complementary textual and spatial priors. Motivated by our observation that the minimal enclosing bounding box derived from RES rapidly achieves accurate localization during early training, HPC incorporates a global sentence-level feature to capture holistic semantic guidance and adaptively fuses the mask-based coordinate as an explicit reference into the query and its position embedding. This design enables REC to benefit from the spatial knowledge acquired through dense segmentation, while the localization objective in turn provides geometry constraints for refining the segmentation results. Overall, DeCo establishes a novel MTVG paradigm that first alleviates task interference through representation decoupling and then exploits task complementarity through prior coupling.

Extensive experiments on RefCOCO/+, G-Ref, ReferIt, Flickr, DIOR-RSVG, RRSIS-D, and other benchmarks demonstrate the effectiveness and generalization of our method. As shown in \cref{fig_overview} (e), DeCo consistently surpasses MLLM-based SOTA method~\citep{text4seg++} on both REC and RES, improving the average performance by 1.1\% and 3.1\%, respectively. Moreover, \cref{fig_overview} (f) illustrates that DeCo achieves significant improvements over task-specific SOTA methods~\citep{bare, wimfris} under both standard and mixed settings, with average gains of 3.6\%/4.4\% on REC and 6.8\%/5.4\% on RES, respectively. In summary, our main contributions are as follows:

\begin{itemize}
    \item We identify and revisit the representation-sharing paradigm for MTVG, showing that REC and RES exhibit distinct visual preferences and that simply jointly optimizing them over shared representations is insufficient to fully exploit their complementarity.
    \item We propose DeCo, an efficient Decouple-to-Couple learning framework, which first separates task-oriented representation learning through Task-aware Semantic Decoupling (TSD) and subsequently exploits complementary semantic and spatial priors through Hybrid Prior Coupling (HPC), providing a simple and effective paradigm for MTVG.
    \item We establish strong empirical results with parameter-efficient adaptation, where DeCo introduces a small number of trainable parameters and achieves state-of-the-art performance across multiple REC and RES benchmarks, consistently outperforming both MLLM-based and task-specific competitors under different evaluation settings.
\end{itemize}

%% file: Sections/2_related_work.tex
\section{Related Work}

\paragraph{Multi-Task Visual Grounding} 
Multi-task visual grounding (MTVG)~\citep{mcn} aims to jointly identify a language-referred object at both box- and pixel-level granularity. Early MTVG methods typically leverage separate visual and language encoders, such as Swin Transformer~\citep{swin} and BERT~\citep{bert}, to construct multimodal representations. These methods design multiple task-specific heads to perform REC and RES in parallel~\citep{reftr, vglaw, eevg}, or formulate both objectives in a unified point-sequence regression framework~\citep{seqtr, polyformer, pvd}. With the emergence of vision-language pre-trained models, subsequent works~\citep{propvg, latent, gc3vg} increasingly explore a unified multimodal encoder, such as CLIP~\citep{clip} and BEiT3~\citep{beit3}. OneRef~\citep{oneref} introduces a mask referring modeling paradigm to establish a unified visual-language feature space. C$^3$VG~\citep{c3vg} adopts a coarse-to-fine framework that uses bidirectional consistency constraints to improve the agreement between REC and RES. InstanceVG~\citep{instancevg} further employs instance queries with reference points to jointly predict instance-level boxes and masks. However, these methods generally require full tuning, resulting in considerable optimization costs. More recently, the rapid progress of MLLMs has motivated researchers to exploit their strong reasoning and generalization capabilities~\citep{llava, internvl2.5} for MTVG. These methods typically perform REC through coordinate regression, while addressing RES via mask generation~\citep{sam, unipixel, fclm}, contour prediction~\citep{visionllm, vistallm, visionllmv2}, or semantic correspondence~\citep{text4seg, text4seg++}. Although these methods reduce the training cost through parameter-efficient adaptation~\citep{lora, adapter} or instruction tuning~\citep{instruction, lora2}, inference remains computationally intensive due to billions of parameters. More importantly, all the aforementioned methods adopt a shared-representation paradigm, overlooking the distinct feature preferences of REC and RES. In contrast, DeCo explicitly separates representation learning for the two objectives and subsequently exploits their complementary priors. By combining low-rank adaptation (LoRA) with lightweight multi-task heads, DeCo achieves state-of-the-art performance with modest computational overhead.

\paragraph{Parameter-Efficient Fine-Tuning} 
Parameter-efficient fine-tuning (PEFT) aims to transfer frozen large foundation models to various downstream tasks while substantially reducing training costs~\citep{peft_survey1, peft_survey2}. Representative PEFT approaches, including adapter tuning~\citep{adapter, adaptformer}, prompt tuning~\citep{vpt, qformer}, and low-rank reparameterization~\citep{lora, lorand}, have achieved remarkable success in visual perception~\citep{mona, freqs, dsgcr}, language understanding~\citep{pica, perl, hsplitlora}, and multimodal reasoning~\citep{vgent, memba, UGround}. Among them, LoRA~\citep{lora} is one of the most widely adopted and effective methods, which introduces low-rank trainable matrices to parameterize task-specific updates. Recent studies have investigated PEFT for REC~\citep{hivg, swimvg, bare} and RES~\citep{barleria, detris, wimfris} individually, demonstrating competitive performance against fully trained methods. For REC, BARE~\citep{bare} enhances referential understanding by mitigating multimodal bias and strengthening semantic reasoning, while WIMFRIS~\citep{wimfris} improves RES through hierarchical multi-scale feature fusion and lightweight vision-language alignment. However, the effectiveness of PEFT for MTVG remains largely unexplored. DeCo bridges this gap by combining LoRA with our decouple-to-couple learning paradigm, where LoRA serves as a fundamental technique for enabling efficient MTVG adaptation rather than the methodological focus of this work.

%% file: Sections/3_methodology.tex
\section{Methodology}

\subsection{Overview}
\label{sec_overview}

Given an image $I \in \mathbb{R}^{h \times w \times 3}$ and a linguistic expression $T$, MTVG aims to predict the corresponding bounding box $\mathbf{B}$ and segmentation mask $\mathbf{M}$. We employ BEiT3~\citep{beit3} as the frozen multimodal encoder with LoRA~\citep{lora} adaptation to efficiently extract visual patch embeddings $\mathbf{V}=[\bar{\mathbf{v}}, \mathbf{v}_1, \cdots, \mathbf{v}_n]\in \mathbb{R}^{(n+1)\times d}$ and textual token embeddings $\mathbf{T}=[\bar{\mathbf{t}}, \mathbf{t}_1, \cdots, \mathbf{t}_m]\in \mathbb{R}^{(m+1)\times d}$, where $n=hw/p_{size}^2$ and $m$ denote the numbers of visual patches and textual tokens, respectively, $\bar{\mathbf{v}}$ and $\bar{\mathbf{t}}$ represent the class tokens of the two modalities, and $d$ is the embedding dimension. Subsequently, we leverage SimFPN~\citep{vitdet} to project $\mathbf{V}$ into low-dimensional multi-scale features $\mathcal{V}=\{\mathbf{V}_l\}_{l=1}^{L}$, where $\mathbf{V}_l\in\mathbb{R}^{h_l\times w_l \times c}$, $L=4$, and $c<d$. Detailed preliminaries of LoRA and SimFPN are provided in Appendix \ref{sec_app_encoding}. Based on this multimodal encoding architecture, we initially construct three effective baseline models (\textit{i.e.}, LoM, SoM, and DoM) to improve MTVG. As discussed in \cref{sec_introduction} and \cref{fig_overview} (a)-(c), LoM focuses on REC but is weaker on RES, whereas SoM exhibits the opposite behavior. By integrating both over the shared $\mathcal{V}$, DoM inherits the strengths of LoM and SoM, yet with only marginal additional gains. This observation motivates us to rethink the role of shared representations in MTVG. 

To this end, we propose DeCo, which follows a decouple-to-couple learning paradigm. As illustrated in \cref{fig_overview} (d), DeCo first introduces a task-aware semantic decoupling module to route the shared $\mathcal{V}$ into localization- and segmentation-oriented multi-scale features, $\mathcal{V}^{loc}$ and $\mathcal{V}^{seg}$, under the guidance of the salient word-level feature $\mathbf{t}_{s}$. It guarantees that the two objectives learn distinct representations from their respective feature interests while preserving the efficiency of shared multi-scale encoding. Subsequently, DeCo further leverages a hybrid prior coupling module to exploit the complementary information across two tasks. Specifically, a Transformer layer is employed to inject text-based semantic prior into an object query, with the entire sentence-level feature $\mathbf{T}$ serving as the key and value. Then, a mask-derived spatial prior is obtained from $\mathbf{M}$ through a mask-to-box strategy and adaptively incorporated into the query and its position embedding via a learnable gate. This design enables REC to benefit from the spatial knowledge acquired through dense segmentation, while the localization objective in turn provides geometric constraints for rectifying the result of RES.

\subsection{Task-aware Semantic Decoupling}

The widely adopted shared-representation paradigm overlooks the distinctiveness of REC and RES. To capture discriminative features for each objective, we propose Task-aware Semantic Decoupling (TSD), which decouples $\mathcal{V}$ via salient word-level modulation and lightweight task-specific experts.

\paragraph{Salient Word-level Modulation.} 
A referred object is typically characterized by a subset of semantic attributes in the referring expression. Thus, we initially identify a salient word-level feature from the textual embeddings $\mathbf{T}=[\bar{\mathbf{t}},\mathbf{t}_1,\ldots,\mathbf{t}_m]$. Specifically, we compute the cosine similarity between each token and the class token $\bar{\mathbf{t}}$, and select the token with the highest similarity as $\mathbf{t}_s$:
\begin{equation}
j^*=
\underset{j}{\arg\max}\ 
\frac{\mathbf{t}_j^{\top}\bar{\mathbf{t}}}
{\|\mathbf{t}_j\|_2\|\bar{\mathbf{t}}\|_2},
\qquad
j \in \{1, \cdots, m\},
\qquad
\mathbf{t}_s=\mathbf{t}_{j^*}\in \mathbb{R}^{1\times d}.
\end{equation}
This selection provides a compact semantic cue that adaptively captures the
most discriminative target-related concept while avoiding the redundancy of the entire sentence representation. For each multi-scale feature $\mathbf{V}_l\in\mathbb{R}^{h_l\times w_l\times c}$, we flatten its spatial dimensions into visual tokens $\mathbf{X}_l\in\mathbb{R}^{h_lw_l\times c}$ and align $\mathbf{t}_s$ into the same feature space through a learnable projection $\mathbf{W}_t \in \mathbb{R}^{d\times c}$. We then adopt a standard cross-attention operator with $\mathbf{t}_s$ as the query and $\mathbf{X}_l$ as the key and value:
\begin{equation}
\mathbf{Q}_l=(\mathbf{t}_s\mathbf{W}_t)\mathbf{W}_q,\qquad
\mathbf{K}_l=\mathbf{X}_l\mathbf{W}_k,\qquad
\mathbf{U}_l=\mathbf{X}_l\mathbf{W}_v,
\end{equation}
\begin{equation}
\mathbf{D}_l=
\operatorname{softmax}
\left(
\frac{\mathbf{Q}_l\mathbf{K}_l^{\top}}
{\sqrt{d_k}}
\right),
\qquad
\mathbf{R}_l=\mathbf{D}_l\mathbf{U}_l,
\end{equation}
where $\mathbf{W}_{q/k/v}\in \mathbb{R}^{c\times c}$ denotes a learnable query$/$key$/$value projection. Here, $\mathbf{D}_l\in\mathbb{R}^{1\times h_lw_l}$ provides a spatial relevance distribution over visual tokens with respect to the target semantic, while $\mathbf{R}_l\in \mathbb{R}^{1\times c}$ aggregates it into a word-conditioned global response.

\paragraph{Task-specific Experts.} 
We exploit the two complementary signals from cross-attention to construct localization- and segmentation-oriented representations. Both experts adopt the same lightweight architecture, consisting of two $1\times1$ convolutional layers interleaved with GELU activation and LayerNorm. For localization, the word-conditioned global response $\mathbf{R}_l$ is broadcastly connected with the original visual tokens, enhancing semantic cues that facilitate object-level correspondence:
\begin{equation}
\mathbf{V}^{loc}_l =
\mathcal{E}^{loc}_l
\big(
\operatorname{reshape}
(
\mathbf{X}_l+\mathbf{R}_l
)
\big),
\end{equation}
where $\mathcal{E}^{loc}_l$ denotes the REC-specific expert network and $\operatorname{reshape}(\cdot)$ restores the spatial dimensions of the visual tokens to $h_l\times w_l$. For segmentation, we instead use the spatial relevance distribution $\mathbf{D}_l$ to reweight the shared visual tokens, highlighting target-relevant regions while preserving their fine-grained spatial structure, which can be mathematically formulated as follows:
\begin{equation}
\mathbf{V}^{seg}_l =
\mathcal{E}^{seg}_l
\big(
\operatorname{reshape}
(
\mathbf{X}_l \odot \mathbf{D}_l^{\top}
)
\big),
\end{equation}
where $\odot$ represents element-wise multiplication and $\mathcal{E}^{seg}_l$ is the RES-specific expert network. This asymmetric design allows localization to incorporate explicit word-aware semantic responses while segmentation focuses on spatially relevant visual regions. The resulting task-specific multi-scale representations $\mathcal{V}^{loc} \in \{\mathbf{V}_l^{loc}\}_{l=1}^{L}$ and $\mathcal{V}^{seg} \in \{\mathbf{V}_l^{seg}\}_{l=1}^{L}$ are subsequently delivered to their respective branches, thereby alleviating the mutual competition arising from shared representations and allowing each task to preserve its preferred visual cues.

\subsection{Hybrid Prior Coupling}

While TSD ensures task-oriented features, localization and segmentation remain inherently complementary. In particular, dense pixel supervision enables RES to produce a reliable object proposal earlier in training than sparse box supervision used in REC. To exploit this property, we introduce Hybrid Prior Coupling (HPC), which integrates sentence-level semantic prior and mask-derived spatial prior into the localization query, thereby establishing a collaborative coupling mechanism.

\paragraph{Sentence-level Semantic Prior.} 
Given the sentence-level feature $\mathbf{T}$, we initially project it into the query embedding space via a linear layer $\mathbf{W}_{\rm T} \in \mathbb{R}^{d\times c}$. A Transformer layer is then employed to incorporate the textual semantic information into the randomly initialized object query $\mathbf{q}\in \mathbb{R}^{1\times c}$. Specifically, the object query first interacts with itself through self-attention and subsequently aggregates global linguistic contexts via cross-attention, followed by a feed-forward network (FFN):
\begin{equation}
\tilde{\mathbf{q}}
=
\operatorname{FFN}
\Big(
\operatorname{CrossAttn}
\big(
\operatorname{SelfAttn}
(
\mathbf{q} + \mathbf{p}
) 
+ \mathbf{p},
\mathbf{T}\mathbf{W}_{\rm T},
\mathbf{T}\mathbf{W}_{\rm T}
\big)
\Big),
\end{equation}
where $\mathbf{p}\in \mathbb{R}^{1\times c}$ denotes the position embedding of $\mathbf{q}$. Unlike TSD, which employs the salient word-level feature to provide a compact discriminative cue for feature decoupling, HPC utilizes the entire sentence-level feature to capture holistic linguistic semantics for object localization, explicitly grounding the initial query in the semantic space of the referring expression.

\paragraph{Mask-derived Spatial Prior.} 
Pixel-level segmentation inherently provides informative object geometry that can complement the semantic query. To further transfer this prior knowledge, we convert the predicted binary mask $\mathbf{M}$ into a normalized bounding box $\mathbf{B}_{m2b}$ through a mask-to-box strategy. Given coordinate grids $x_{ij}$ and $y_{ij}$, we apply each grid with an extremal constant according to the mask value and aggregate the foreground extremes via global min and max reductions as follows:
\begin{equation}
\begin{aligned}
x_{\min} &= \min_{i,j} \big( \mathbf{M}_{ij} x_{ij} + (1-\mathbf{M}_{ij}) w \big), \\
x_{\max} &= \max_{i,j} \big( \mathbf{M}_{ij} x_{ij} - (1-\mathbf{M}_{ij}) \big),
\end{aligned}
\qquad
\begin{aligned}
y_{\min} &= \min_{i,j} \big( \mathbf{M}_{ij} y_{ij} + (1-\mathbf{M}_{ij}) h \big), \\
y_{\max} &= \max_{i,j} \big( \mathbf{M}_{ij} y_{ij} - (1-\mathbf{M}_{ij}) \big),
\end{aligned}
\end{equation}
\begin{equation}
\mathbf{B}_{m2b} = \big[ (x_{\min}+x_{\max}) / 2w,\ (y_{\min}+y_{\max})/ 2h,\ (x_{\max}-x_{\min}) / w,\ (y_{\max}-y_{\min}) / h \big],
\end{equation}
which is further mapped into the query and position embedding spaces, respectively: 
\begin{equation}
\mathbf{q}_{m2b} = \mathbf{B}_{m2b}\mathbf{W}_{m2b}^{q},
\qquad
\mathbf{p}_{m2b} = \mathbf{B}_{m2b}\mathbf{W}_{m2b}^{p},
\end{equation}
where $\mathbf{W}_{m2b}^{q},\mathbf{W}_{m2b}^{p}\in\mathbb{R}^{4\times c}$ are two projection matrices. To adaptively balance the contributions of semantic and spatial information, we design a lightweight gating function conditioned on both the semantic-enriched query and the mask-derived bounding box:
\begin{equation}
\gamma =
\sigma
\Big(
\operatorname{MLP}
\big(
\operatorname{Concat}(\tilde{\mathbf{q}};\mathbf{B}_{m2b})
\big)
\Big),
\end{equation}
where $\operatorname{Concat}(\cdot\,;\cdot)$, $\operatorname{MLP}(\cdot)$, and $\sigma(\cdot)$ represent concatenation, two linear layers separated by a GELU activation, and the sigmoid function, respectively. The gate value $\gamma \in (0, 1)$ is then utilized to get the final hybrid prior query and its position embedding as follows:
\begin{equation}
\hat{\mathbf{q}}
=
\gamma\hspace{0.5pt}\tilde{\mathbf{q}}
+
(1-\gamma)\ \mathbf{q}_{m2b}, \qquad 
\hat{\mathbf{p}} = \gamma\hspace{0.5pt}\mathbf{p} + (1-\gamma)\mathbf{p}_{m2b}.
\end{equation}
In this way, HPC evolves the randomly initialized object query into a strong context-geometric anchor that accelerates the convergence of the localization branch, while the resulting detection loss in turn provides bounding constraints that further refine the segmentation mask.

\subsection{Training Objectives}

Given the hybrid prior query $\hat{\mathbf{q}}$, its position embedding $\hat{\mathbf{p}}$, and REC-specific features $\mathcal{V}^{loc}$, we first employ Deformable DETR~\citep{deformabledetr} to produce the output query $\mathbf{q}_{\rm o}$. Subsequently, a box FFN regresses $\mathbf{q}_{\rm o}$ into the predicted box $\mathbf{B}$, while a mask FFN projects $\mathbf{q}_{\rm o}$ into a mask embedding, which is multiplied with the high-resolution segmentation feature $\mathbf{V}_1^{seg}$ to obtain the corresponding instance mask $\mathbf{M}_{ins}$. In parallel, RES-specific features $\mathcal{V}^{seg}$ are fed into UNet~\citep{unet} to get the predicted mask $\mathbf{M}$. The overall prediction process can be formulated as:
\begin{equation}
    \mathbf{q}_{\rm o} = \operatorname{DETR}(\hat{\mathbf{q}},~\hat{\mathbf{p}},~\mathcal{V}^{loc}),~\mathbf{B} = \operatorname{FFN}_b(\mathbf{q}_{\rm o}),~\mathbf{M}_{ins} = \operatorname{FFN}_m(\mathbf{q}_{\rm o})\otimes\mathbf{V}_1^{seg},~\mathbf{M} = \operatorname{UNet}(\mathcal{V}^{seg})
\end{equation}
where $\otimes$ denotes Einstein multiplication. For box regression, we adopt L1 loss and generalized IoU loss, while a combination of binary cross-entropy and Dice losses is used for mask prediction:
\begin{equation}
\begin{aligned}
\mathcal{L}_{det} &= \mathcal{L}_{l1}(\mathbf{B}, \mathbf{B}^*) + \mathcal{L}_{giou}(\mathbf{B}, \mathbf{B}^*), \\
\mathcal{L}_{seg} &= \mathcal{L}_{bce}(\mathbf{M}, \mathbf{M}^*) + \mathcal{L}_{dice}(\mathbf{M}, \mathbf{M}^*),
\end{aligned}
\quad
\begin{aligned}
\mathcal{L}_{ins} &= \mathcal{L}_{bce}(\mathbf{M}_{ins}, \mathbf{M}^*) + \mathcal{L}_{dice}(\mathbf{M}_{ins}, \mathbf{M}^*), \\
\mathcal{L}_{m2b} &= \mathcal{L}_{l1}(\mathbf{B}_{m2b}, \mathbf{B}^*) + \mathcal{L}_{giou}(\mathbf{B}_{m2b}, \mathbf{B}^*),
\end{aligned}
\end{equation}
where $\mathbf{B}^*$ and $\mathbf{M}^*$ are the ground-truth bounding box and segmentation mask. Finally, the total training objective is defined as a simple weighted sum of the four loss functions with same weights:
\begin{equation}
\mathcal{L}_{total} = \mathcal{L}_{det} + \mathcal{L}_{ins} + \mathcal{L}_{seg} + \mathcal{L}_{m2b}.
\end{equation}
During inference, only $\mathbf{B}$ and $\mathbf{M}$ are retained as the final REC and RES predictions. The auxiliary outputs $\mathbf{M}_{ins}$ and $\mathbf{B}_{m2b}$ are involved in the forward process but are discarded from the final outputs, requiring no additional post-processing.

%% file: Sections/4_experiments.tex
\input{Tables/1_rec}

\input{Tables/2_res}

\section{Experiments}

\subsection{Experimental Setup}

\textbf{Datasets \& Metrics.} 
We conduct extensive experiments on ten benchmarks covering both natural and remote sensing images. For natural images, we evaluate DeCo on \textbf{RefCOCO}~\citep{refcoco}, \textbf{RefCOCO+}~\citep{refcoco}, and \textbf{G-Ref}~\citep{gref2, gref1} for both REC and RES, while \textbf{ReferIt}~\citep{referit} and \textbf{Flickr}~\citep{flickr30k} for REC. To further examine cross-domain generalization, we additionally evaluate DeCo on five remote sensing benchmarks, including \textbf{DIOR-RSVG}~\citep{mgvlf}, \textbf{SARVG1.0}~\citep{tacmt}, \textbf{RRSIS-D}~\citep{rmsin}, \textbf{RIS-LAD}~\citep{ris_lad}, and \textbf{RefDIOR}~\citep{rrsecs}. Following previous works~\citep{oneref, instancevg}, we report P@.5 for REC, while oIoU and mIoU are for RES. In the ablation studies, we additionally report P@.5:.95 for a more precise REC evaluation. More descriptions about datasets and metrics are provided in Appendix~\ref{sec_app_dataset} and \ref{sec_app_metric}.

\textbf{Implementation\,Details.} 
We adopt two model scales, BEiT3-B and BEiT3-L~\citep{beit3}, as frozen multimodal encoders and introduce LoRA~\citep{lora} for parameter-efficient adaptation. For most experiments, we train DeCo using the Adam optimizer for $35$ epochs, with the learning rate reduced by a factor of $10$ at the $28$-th epoch. The input image is resized to $448\times 448$, and the maximum expression length is limited to $20$ tokens. The initial learning rate is set to $1\times10^{-4}$, with the batch size of $32$. More implementation details are provided in Appendix~\ref{sec_app_implem}.

\subsection{Main Results}

\textbf{Referring Expression Comprehension.} 
As evaluated in \cref{tab_rec} and \cref{tab_other_rec}, DeCo consistently achieves state-of-the-art performance compared to all MLLM-based and task-specific methods across both standard and mixed settings. With frozen BEiT3-B/L~\citep{beit3} backbones, DeCo sets top results on all eight splits under the mixed setting (90.5\%/93.1\% average P@.5) and delivers gains of up to 3.9\%, 3.3\%, and 0.8\% on G-Ref (val-g), ReferIt, and Flickr, respectively. These results validate that decoupling task-preferred representations and coupling complementary priors effectively eliminates task interference while generalizing across diverse domains.

\textbf{Referring Expression Segmentation.} 
As shown in \cref{tab_res} and \cref{tab_other_res}, DeCo sets new RES records across all benchmarks under the standard setting. With BEiT3-B/L, DeCo outperforms OneRef by 2.1\%/2.9\% average oIoU while keeping backbones frozen. Compared with MLLM-based methods, DeCo surpasses Text4Seg++-14B by 3.1\% average oIoU and the strongest competitor STAMP-7B on all eight splits. Furthermore, DeCo exceeds DETRIS and WIMFRIS by significant margins of 13.4\% oIoU and 9.8\% mIoU on G-Ref (val-g), highlighting robust generalization to complex expressions. More RES results with mIoU and other remote sensing experiments are listed in Appendix~\ref{sec_app_exps}.

\subsection{Ablation Studies}

\input{Tables/3_other}

\textbf{Effects of Core Components.} 
We first investigate the contribution of the core components of DeCo, including LoM, SoM, TSD, and HPC, as reported in \cref{tab_abla_comp}. Starting from the single-task baselines, LoM and SoM achieve average scores of 74.0\% and 73.9\%, respectively, while integrating them in DoM shows limited gains from naive joint optimization. Adding TSD or HPC individually improves both objectives, confirming that task decoupling and prior coupling address distinct interference sources. Combining both, DeCo achieves the best average performance of 76.6\%, exceeding DoM by 2.2\%. Qualitative analysis in \cref{fig_comp} intuitively confirms the role of each component.

\textbf{Effects of PEFT and LoRA $r$ Settings.} 
We next examine the efficiency of PEFT the impact of the LoRA rank $r$ in \cref{tab_abla_peft}. Full tuning yields only 61.4\% despite 221.3M trainable parameters and 81.0G FLOPs, whereas freezing drops to 54.1\%, highlighting the necessity of lightweight adaptation. In contrast, LoRA provides a substantially more favorable accuracy-efficiency trade-off. With only 0.6M trainable parameters and 0.3G FLOPs at $r=4$, the model already achieves an average score of 75.0\% and rises steadily to 76.6\% at $r=32$, which is set to the trade-off setting.

\textbf{Effects of LoRA Insertion Positions.} 
As listed in \cref{tab_abla_pos}, query-only adaptation yields only 72.9\% average performance, while value and query-value adaptation rise to 75.0\% and 75.8\%, suggesting important roles of information retrieval and feature transformation in effective multimodal representation adaptation. Adapting the complete attention block (Q-K-V-O) further improves the average score to 76.6\%, providing a favorable balance between performance and computational overhead.

\textbf{Effects of Segmentation Heads.} 
We further compare different segmentation heads, including UNet~\citep{unet}, SegFormer~\citep{segformer}, UPerNet~\citep{upernet}, ReMamber~\citep{remamber}, LAVT-RS~\citep{lavt_rs}, and DETRIS~\citep{detris}, as summarized in \cref{tab_abla_seg}. The lightweight UNet achieves the best average performance with only a 0.4M/2.6G burden, while other classic or multimodal segmentors offer no gains at much higher costs, indicating the efficiency primarily comes from our decouple-to-couple learning rather than a complex decoder. More ablation studies and qualitative results are provided in Appendix~\ref{sec_app_abla}.

%% file: Tables/1_rec.tex
\begin{table*}[t]
\centering

\caption{Main results of \textbf{REC} models using \textbf{P@.5} on RefCOCO, RefCOCO+, and G-Ref datasets. The best results are highlighted in \textbf{bold}, and the second-best are \underline{underlined}. $^{*}$ indicates that the backbone is frozen with only the additional introduced adaptation parameters being optimized.}
\label{tab_rec}
\vspace{-0.5em}
\setlength{\tabcolsep}{4pt}
\resizebox{\linewidth}{!}{
\begin{tabular}{l|c|ccc|ccc|cc|c}
\toprule
\multirow{2}{*}{Models} & Vision\,/\,Language & \multicolumn{3}{c|}{RefCOCO} & \multicolumn{3}{c|}{RefCOCO+} & \multicolumn{2}{c|}{G-Ref} & \multirow{2}{*}{Avg.} \\  \cmidrule{3-10}
 &  Backbone  & val     & testA   & testB   & val+     & testA+    & testB+   & val-u        & test-u   &     \\
                        
\midrule
\multicolumn{11}{l}{\textit{MLLM-based Methods}}  \\
\midrule

GSVA-13B~\citep{gsva}    & SAM-H\,/\,Llama2$^*$              & 89.2    & 92.1    & 87.2    & 79.7    & 84.5     & 73.4    & 85.5         & 86.2          & 84.7                 \\
LION-12B~\citep{lion}    & CLIP-G$^*$\,/\,FlanT5-XXL$^*$         & 89.8          & 93.0          & 85.6          & 84.0          & 89.2          & 78.1          & 85.5          & 85.7          & 86.4                 \\
Groma-7B~\citep{groma}     & DINOv2-L$^*$\,/\,Vicuna-v1.5      & 89.5    & 92.1    & 86.3    & 83.9    & 88.9     & 78.1    & 86.4         & 87.0          & 86.5                 \\
SegLLM-7B~\citep{segllm}    & CLIP-L$^*$\,/\,Vicuna-v1.5     & 90.0    & 92.1    & 86.2    & 82.2    & 85.5     & 76.1    & 83.9         & 85.9          & 85.2                 \\
Text4Seg-13B~\citep{text4seg}     & SAM-H\,/\,LLaVa-v1.5$^*$          & 91.2    & 94.3    & 88.0    & 85.7    & 90.8     & 80.1    & 85.6         & 85.5          & 87.7                 \\
UniPixel-7B~\citep{unipixel}      & SAM2-B+\,/\,Qwen2.5-VL$^*$               & 92.0    & 94.4    & 88.1    & 87.2    & 91.9     & 82.1    & 88.6         & 88.7          & 89.1                 \\
UFO-8B~\citep{ufo}     & InternViT$^*$\,/\,InternLM2.5             & 93.1          & 94.8          & 89.2          & 87.7          & 92.1          & 82.3          & 88.2          & 89.2          & 89.6                 \\
VPP-7B~\citep{vpp}     & CLIP-L\,/\,Vicuna-v1.5         & 90.4    & 92.9    & 85.8    & 84.7    & 89.8     & 77.0    & 85.3         & 85.5          & 86.4                 \\
SIFThinker-7B~\citep{sifthinker}      & Qwen2.5-VL\,/\,Qwen2.5-VL     & \underline{93.8}    & \underline{95.1}    & \underline{90.4}    & 86.0    & 90.7     & 80.5    & \underline{90.4}     & \underline{90.6}   & 89.7      \\
VGent-16B~\citep{vgent}     & Qwen2.5-VL\,/\,Qwen2.5-VL$^*$    & 92.4    & 94.7    & 89.8    & \underline{88.1}    & \underline{92.2}     & \underline{83.3}    & 90.4         & 90.1    & \underline{90.1}                 \\
Text4Seg++-14B~\citep{text4seg++}     & SAM-H\,/\,Qwen2.5-VL$^*$            & \textbf{94.3}    & \textbf{96.2}    & \textbf{91.4}    & \textbf{90.6}    & \textbf{94.2}     & \textbf{86.1}    & \textbf{91.6}   & \textbf{91.8}    & \textbf{92.0}       \\

\midrule
\multicolumn{11}{l}{\textit{Task-specific Methods (Standard Setting)}}  \\
\midrule

TransVG++~\citep{transvg++}  & ViTDet-B\,/\,BERT-B & 86.3      & 88.4      & 81.0      & 75.4      & 80.5      & 66.3      & 76.2      & 76.3      & 78.8      \\
EEVG~\citep{eevg}       & ViTDet-B\,/\,BERT-B & 88.1      & 90.3      & 85.5      & 78.0      & 82.4      & 69.2      & 79.6      & 80.2      & 81.7      \\
HiVG~\citep{hivg}       & CLIP-B$^*$\,/\,CLIP-B$^*$          & 87.3      & 89.9      & 83.3      & 78.1      & 83.8      & 68.1      & 78.3      & 78.8      & 80.9      \\
SimVG~\citep{simvg}      & BEiT3-B\,/\,BEiT3-B         & 87.6      & 90.2      & 84.0      & 78.7      & 83.4      & 71.8      & 80.4      & 80.5      & 82.1      \\
OneRef~\citep{oneref}     & BEiT3-B\,/\,BEiT3-B         & 88.8      & 91.0      & 85.3      & 80.4      & 86.5      & 74.3      & \underline{83.7}      & 83.5      & 84.2      \\
PropVG~\citep{propvg}     & BEiT3-B\,/\,BEiT3-B         & 89.0      & 91.6      & 85.7      & \underline{83.7}      & \underline{88.0}      & \underline{76.6}      & 83.5      & \underline{84.4}      & \underline{85.3}      \\
Latent-VG~\citep{latent}  & BEiT3-B\,/\,BEiT3-B         & \underline{89.8}      & \underline{92.4}      & \underline{85.9}      & 83.2      & 87.7      & 76.5      & 82.6      & 83.1      & 85.1      \\
MAP~\citep{map}        & DINOv2-B$^*$\,/\,BERT-B$^*$ & 86.9      & 89.4      & 82.0      & 75.7      & 82.6      & 66.4      & 79.7      & 79.3      & 80.2      \\
SwimVG~\citep{swimvg}     & DINOv2-B$^*$\,/\,CLIP-B$^*$ & 88.3      & 90.4      & 84.9      & 77.9      & 83.2      & 70.0      & 79.1      & 80.1      & 81.7      \\
OneSVG~\citep{onesvg}     & BEiT3-B$^*$\,/\,BEiT3-B$^*$       & 88.8      & 91.0      & 85.6      & 81.7      & 86.6      & 75.6      & 82.4      & 82.1      & 84.2      \\
BARE~\citep{bare}     & BEiT3-B$^*$\,/\,BEiT3-B$^*$       & 88.7 & 91.6 & 84.8 & 81.3 & 86.8 & 74.8 & 82.7 & 83.2 & 84.2 \\
\rowcolor[HTML]{ECECEC}
\textbf{DeCo (Ours)} & BEiT3-B$^*$\,/\,BEiT3-B$^*$    & \textbf{90.9} & \textbf{93.2} & \textbf{87.3} & \textbf{86.4} & \textbf{90.7} & \textbf{80.2} & \textbf{86.4} & \textbf{87.0} & \textbf{87.8} \\
\midrule

HiVG~\citep{hivg}       & CLIP-L$^*$\,/\,CLIP-L$^*$          & 88.1      & 91.1      & 83.7      & 80.1      & 86.8      & 70.5      & 80.8      & 80.3      & 82.7      \\
SimVG~\citep{simvg}      & BEiT3-L\,/\,BEiT3-L         & 90.6      & 92.5      & 87.7      & 85.4      & 89.6      & 79.7      & 86.0      & 86.8      & 87.3      \\
OneRef~\citep{oneref}     & BEiT3-L\,/\,BEiT3-L         & \underline{92.9}      & \underline{94.0}    & \underline{90.2}      & \underline{88.0}      & 91.6      & \underline{83.7}      & 88.1      & \underline{89.3}      & \underline{89.7}      \\
OneSVG~\citep{onesvg}     & BEiT3-L$^*$\,/\,BEiT3-L$^*$         & 91.1      & 93.3      & 88.3      & 87.4      & 90.9      & 81.2      & 87.5      & 87.6      & 88.4      \\
BARE~\citep{bare}     & BEiT3-L$^*$\,/\,BEiT3-L$^*$      & 92.4 & 93.8 & 89.2 & 87.9 & \underline{91.8} & 83.2 & \underline{88.8} & 88.9 & 89.5 \\
\rowcolor[HTML]{ECECEC}
\textbf{DeCo (Ours)} & BEiT3-L$^*$\,/\,BEiT3-L$^*$   & \textbf{93.8} & \textbf{95.1} & \textbf{91.1} & \textbf{90.6} & \textbf{94.1} & \textbf{85.1} & \textbf{90.5} & \textbf{91.0} & \textbf{91.4}  \\

\midrule
\multicolumn{11}{l}{\textit{Task-specific Methods (Mixed Setting)}}  \\
\midrule

PolyFormer~\citep{polyformer}       & Swin-B\,/\,BERT-B     & 89.7    & 91.7    & 86.0    & 83.7    & 88.6     & 76.4    & 84.5         & 85.0          & 85.7      \\
EEVG~\citep{eevg}        & ViTDet-B\,/\,BERT-B           & 90.5          & 92.7          & 87.7          & 81.8          & 87.8          & 74.9          & 85.2          & 84.7          & 85.7                 \\
HiVG~\citep{hivg}            & CLIP-B$^*$\,/\,CLIP-B$^*$                    & 90.6    & 92.6    & 87.2    & 83.1    & 89.2     & 76.7    & 84.5         & 85.6          & 86.2                 \\
SimVG~\citep{simvg}          & BEiT3-B\,/\,BEiT3-B                   & 91.5    & 93.7    & 87.9    & 84.8    & 88.9     & 79.1    & 86.3         & 87.3          & 87.4                 \\
OneRef~\citep{oneref}      & BEiT3-B\,/\,BEiT3-B                   & 91.9    & 94.3    & 88.6    & 86.4    & 90.4     & 79.5    & 86.8         & 87.3          & 88.2                 \\
C$^3$VG~\citep{c3vg}      & BEiT3-B\,/\,BEiT3-B                   & 92.5    & 94.6    & 88.7    & \underline{87.4}    & 90.7     & \underline{81.4}    & 87.7         & \underline{88.3}          & 88.9                 \\
PropVG~\citep{propvg}         & BEiT3-B\,/\,BEiT3-B                   & \underline{92.7}    & 95.1    & \underline{89.6}    & 87.3    & \underline{90.9}     & 81.3    & \underline{88.2}         & 88.3          & \underline{89.2}                 \\
Latent-VG~\citep{latent}     & BEiT3-B\,/\,BEiT3-B                   & 91.8    & 94.6    & 88.6    & 86.4    & 90.6     & 80.6    & 87.0         & 87.1          & 88.3                 \\
MAP~\citep{map}     & BEiT3-B$^*$\,/\,BEiT3-B$^*$                   & 91.3    & 93.7    & 88.3    & 85.0    & 88.4     & 78.3    & 86.5         & 86.6          & 87.3                 \\
InstanceVG~\citep{instancevg}       & BEiT3-B\,/\,BEiT3-B                   & 92.4    & \underline{95.1}    & 88.8    & 87.1    & 90.8     & 80.7    & 87.2         & 88.1          & 88.8                 \\
BARE~\citep{bare}     & BEiT3-B$^*$\,/\,BEiT3-B$^*$ & 90.3 & 91.5 & 87.9 & 83.1 & 87.1 & 77.3 & 85.4 & 86.0 & 86.1 \\
\rowcolor[HTML]{ECECEC}
\textbf{DeCo (Ours)}          & BEiT3-B$^*$\,/\,BEiT3-B$^*$                  & \textbf{93.4} & \textbf{95.6} & \textbf{90.5} & \textbf{89.4} & \textbf{92.7} & \textbf{83.7} & \textbf{89.2} & \textbf{89.8} & \textbf{90.5}  \\

\midrule

PolyFormer~\citep{polyformer}       & Swin-L\,/\,BERT-B      & 90.4    & 92.9    & 87.2    & 85.0    & 89.8     & 78.0    & 85.8    & 85.9     & 86.9     \\
UNINEXT~\citep{uninext}           & ConvNeXt-L\,/\,BERT-B           & 91.4          & 93.7          & 88.9          & 83.1          & 87.9          & 76.2          & 86.9          & 87.5          & 87.0                 \\
Grounding-DINO~\citep{grounding}       & Swin-L\,/\,BERT-B             & 90.6    & 93.2    & 88.2    & 82.8    & 89.0     & 75.9    & 86.1         & 87.0          & 86.6                 \\
HiVG~\citep{hivg}         & CLIP-L$^*$\,/\,CLIP-L$^*$                    & 90.8    & 92.9    & 88.0    & 86.8    & 89.9     & 78.0    & 86.6         & 86.6          & 87.5                 \\
SimVG~\citep{simvg}       & BEiT3-L\,/\,BEiT3-L                  & 92.9    & 94.7    & 90.3    & 87.3    & 91.6     & 82.4    & 88.0         & 89.2          & 89.6                 \\
OneRef~\citep{oneref}        & BEiT3-L\,/\,BEiT3-L                   & 93.2    & 95.4    & 90.1    & 88.4    & 92.1     & 82.7    & 87.8         & 88.8          & 89.8                 \\
InstanceVG~\citep{instancevg}       & BEiT3-L\,/\,BEiT3-L                   & \underline{94.4}    & \underline{96.0}    & \underline{92.4}    & \underline{90.1}    & \underline{92.9}     & \underline{85.9}    & \underline{89.6}         & \underline{90.6}          & \underline{91.5}                 \\
BARE~\citep{bare}     & BEiT3-L$^*$\,/\,BEiT3-L$^*$ & 92.8 & 94.3 & 90.8 & 88.4 & 91.0 & 84.2 & 90.6 & 90.7 & 90.3 \\
\rowcolor[HTML]{ECECEC}
\textbf{DeCo (Ours)}        & BEiT3-L$^*$\,/\,BEiT3-L$^*$   & \textbf{95.3} & \textbf{96.6} & \textbf{92.7} & \textbf{92.3} & \textbf{95.1} & \textbf{87.9} & \textbf{92.1} & \textbf{92.7} & \textbf{93.1}          \\
\bottomrule
\end{tabular}}
\vspace{-10pt}
\end{table*}

%% file: Tables/2_res.tex
\begin{table*}[t]
\centering

\caption{Main results of \textbf{RES} models using \textbf{oIoU} on RefCOCO, RefCOCO+, and G-Ref datasets.}
\label{tab_res}
\vspace{-0.5em}
\setlength{\tabcolsep}{4pt}
\resizebox{\linewidth}{!}{
\begin{tabular}{l|c|ccc|ccc|cc|c}
\toprule
\multirow{2}{*}{Models} & Vision\,/\,Language & \multicolumn{3}{c|}{RefCOCO}         & \multicolumn{3}{c|}{RefCOCO+}   & \multicolumn{2}{c|}{G-Ref}  & \multirow{2}{*}{Avg.} \\ \cmidrule{3-10}
 & Backbone   & val           & testA         & testB         & val+           & testA+         & testB+  & val-u      & test-u        &               \\
\midrule
\multicolumn{11}{l}{\textit{MLLM-based Methods}}   \\
\midrule

LISA-13B~\citep{lisa}  & SAM-H\,/\,Llama2$^*$   & 76.3 & 78.7 & 72.4 & 66.2 & 71.0 & 59.3 & 70.1 & 71.1 & 70.6 \\ 
GSVA-13B~\citep{gsva}  & SAM-H\,/\,Llama2$^*$  & 79.2 & 81.7 & 77.1 & 70.3 & 73.8 & 63.6 & 75.7 & 77.0 & 74.8 \\
GLaMM-7B~\citep{glamm}        & SAM-H\,/\,LLaVa-v1.5$^*$          & 79.5          & 83.2          & 76.9          & 72.6          & 78.7          & 64.6          & 74.2          & 74.9          & 75.6                 \\
SegLLM-7B~\citep{segllm}     & CLIP-L$^*$\,/\,Vicuna-v1.5      & 80.2          & 81.5          & 75.4          & 70.3          & 73.0          & 62.5          & 72.6          & 73.6          & 73.6                 \\
Text4Seg-13B~\citep{text4seg}        & SAM-H\,/\,LLaVa-v1.5$^*$          & 80.2          & 82.7          & 77.3          & 73.7          & 78.6          & 67.6          & 74.0          & 75.1          & 76.2                 \\
UniPixel-7B~\citep{unipixel}       & SAM2-B+\,/\,Qwen2.5-VL$^*$         & 80.8          & 83.0          & 77.4          & 75.3          & 80.1          & 70.0          & 76.4          & 77.1          & 77.5                 \\
UFO-8B~\citep{ufo}         & InternViT$^*$\,/\,InternLM2.5               & 81.0          & 82.6          & 78.6          & 77.1          & 80.4          & 72.6          & 76.7          & 77.3          & 78.3                 \\
FCLM-7B~\citep{fclm}       & SAM-H$^*$\,/\,Qwen2.5-VL          & \underline{82.6}          & 83.9          & \underline{79.9}          & 77.4         & 80.5          & 71.3          & \underline{79.4}          & \textbf{80.5} & 79.4                 \\
STAMP-7B~\citep{stamp}     & SAM-H$^*$\,/\,Qwen2-VL$^*$            & \textbf{83.1} & \underline{84.5} & \textbf{80.8} & \textbf{79.4} & \textbf{82.8} & \textbf{74.6} & \textbf{79.9} & \underline{80.4}          & \textbf{80.7}        \\
UGround-7B~\citep{UGround}       & SAM-H\,/\,LLaVA-v1.5$^*$          & 80.6          & 83.5          & 77.7          & 72.8          & 77.5          & 65.6          & 74.7          & 76.1          & 76.1                 \\
Text4Seg++-14B~\citep{text4seg++}     & SAM-H\,/\,Qwen2.5-VL$^*$            & 82.5    & \textbf{84.9}    & 79.5    & \underline{77.9}    & \underline{82.5}     & \underline{72.6}    & 79.4   & 79.7    & \underline{79.9}       \\

\midrule
\multicolumn{11}{l}{\textit{Task-specific Methods (Standard Setting)}}   \\
\midrule

MagNet~\citep{magnet}     & Swin-B\,/\,BERT-B   & 75.2          & 78.2          & 71.1          & 66.2          & 71.3          & 58.1          & 65.4          & 66.0          & 68.9          \\
ReMamber~\citep{remamber}   & VMamba-B\,/\,BERT-B & 74.5          & 76.7          & 70.9          & 65.0          & 70.8          & 57.5          & 63.9          & 64.0          & 67.9          \\
BarLeRIa~\citep{barleria} & CLIP-B$^*$\,/\,CLIP-B$^*$   & 75.0    & 77.1   & 71.2          & 68.6   & 73.2     & 61.2       & 65.9     & 66.4      & 69.8          \\
OneRef~\citep{oneref}     & BEiT3-B\,/\,BEiT3-B         & \underline{77.6}    & \underline{81.0} & 73.5          & \underline{70.8}    & \underline{74.5}          & \underline{64.1}    & \underline{70.7}    & \underline{70.6}    & \underline{72.8}    \\
CMANet~\citep{cmanet}     & Swin-B\,/\,BERT-B   & 75.5          & 78.3          & 71.8          & 66.7          & 71.9          & 59.0          & 66.6          & 66.7          & 69.6      \\
MaskRIS~\citep{maskris}    & Swin-B\,/\,BERT-B   & 76.5          & 79.0          & \underline{74.0}    & 67.5          & 74.5          & 59.4          & 65.6          & 66.5      & 70.4   \\
CMIRNet~\citep{cmirnet}  & Swin-B\,/\,BERT-B   & 74.0 & 77.1 & 71.3 & 64.2 & 70.4 & 56.4 & 63.5 & 63.9 & 67.6 \\
DETRIS~\citep{detris}     & DINOv2-B$^*$\,/\,CLIP-B$^*$ & 74.3          & 77.6          & 71.4          & 65.6          & 72.0          & 56.5          & 65.2          & 66.4          & 68.6          \\
PropVG~\citep{propvg}    & BEiT3-B\,/\,BEiT3-B      & 76.8          & 79.6          & 73.7          & 70.2          & 74.3     & 63.4          & 69.3          & 70.5          & 72.2          \\
SaFiRe~\citep{safire}     & Swin-B\,/\,BERT-B   & 75.3          & 78.5          & 71.4          & 66.4          & 72.0          & 58.5          & 67.1          & 66.9          & 69.5          \\
AMLRIS~\citep{amlris}     & Swin-B\,/\,BERT-B   & 75.5          & 78.3          & 72.1          & 67.4          & 73.2          & 59.5       & 65.7          & 66.8          & 69.8          \\
\rowcolor[HTML]{ECECEC}
\textbf{DeCo (Ours)} & BEiT3-B$^*$\,/\,BEiT3-B$^*$         & \textbf{79.0} & \textbf{81.3}    & \textbf{75.7} & \textbf{73.1} & \textbf{77.5} & \textbf{66.3} & \textbf{73.4} & \textbf{73.0} & \textbf{74.9} \\
\midrule

BarLeRIa~\citep{barleria} & CLIP-L$^*$\,/\,CLIP-L$^*$   & 76.8    & 79.0   & 74.0          & 71.5   & 76.2     & 65.4       & 68.7     & 69.7      & 72.7          \\
OneRef~\citep{oneref}     & BEiT3-L\,/\,BEiT3-L     & \underline{80.5}          & \underline{82.8}          & \underline{78.3}          & \underline{74.3}    & \underline{78.4}   & \underline{69.9}   & \underline{74.9}    & \underline{77.4}          & \underline{77.0}          \\
DETRIS~\citep{detris}   & DINOv2-L$^*$\,/\,CLIP-B$^*$ & 76.1          & 78.3     & 73.2          & 67.9     & 73.3          & 60.2          & 66.8     & 68.0    & 70.5          \\
\rowcolor[HTML]{ECECEC}
\textbf{DeCo (Ours)} & BEiT3-L$^*$\,/\,BEiT3-L$^*$    & \textbf{82.7} & \textbf{84.4} & \textbf{80.2} & \textbf{79.2} & \textbf{82.9} & \textbf{73.0} & \textbf{77.9} & \textbf{78.7} & \textbf{79.9}   \\   

\midrule
\multicolumn{11}{l}{\textit{Task-specific Methods (Mixed Setting)}}   \\
\midrule

PolyFormer~\citep{polyformer}        & Swin-B\,/\,BERT-B             & 74.8          & 76.6          & 71.1          & 67.6          & 72.9          & 59.3          & 67.8          & 69.1          & 69.9                 \\
MagNet~\citep{magnet}         & Swin-B\,/\,BERT-B             & 76.6          & 78.3          & 72.2          & 68.1          & 73.6          & 61.8          & 67.8          & 69.3          & 71.0                 \\
BarLeRIa~\citep{barleria} & CLIP-B$^*$\,/\,CLIP-B$^*$   & 77.6    & 79.4   & 75.3          & 71.7   & 75.7     & 66.0       & 70.9     & 71.4      & 73.5          \\
OneRef~\citep{oneref}        & BEiT3-B\,/\,BEiT3-B                   & 81.1          & 83.1          & 77.8          & 72.2          & 77.3          & 67.1          & 75.1          & 77.2          & 76.4                 \\
CMANet~\citep{cmanet}       & Swin-B\,/\,BERT-B             & 77.3          & 80.2          & 73.1          & 68.2          & 73.8          & 60.4          & 67.4          & 70.1          & 71.3                 \\
MaskRIS~\citep{maskris}    & Swin-B\,/\,BERT-B & 78.7          & 80.6          & 75.1          & 70.3          & 75.2          & 62.8          & 69.1          & 71.1          & 72.9          \\
LAVT-RS~\citep{lavt_rs}	&  Swin-B\,/\,BERT-B	&  79.2	&  80.7	&  75.4	&  71.7	&  75.6	&  64.3	&  72.1	&  74.6	&  74.2 \\
CMIRNet~\citep{cmirnet}   & Swin-B\,/\,BERT-B & 77.7 & 79.8 & 74.4 & 68.3 & 74.1 & 61.1 & 69.2 & 71.9 & 72.1 \\
C$^3$VG~\citep{c3vg}       & BEiT3-B\,/\,BEiT3-B                   & 80.9          & 83.2          & 77.9          & 74.7          & 78.0          & 69.0          & 74.4          & 76.4          & 76.8                 \\
PropVG~\citep{propvg}         & BEiT3-B\,/\,BEiT3-B                   & \underline{81.8}          & \underline{83.7}          & \underline{79.3}          & \underline{74.8}          & 78.7          & \underline{69.2}          & 75.5          & \underline{77.4}          & \underline{77.6}                 \\
SaFiRe~\citep{safire}        & Swin-B\,/\,BERT-B             & 81.4          & 83.6          & 78.2          & 74.8          & \underline{78.9}          & 68.3          & \underline{75.8}          & 76.8          & 77.2                 \\
InstanceVG~\citep{instancevg}        & BEiT3-B\,/\,BEiT3-B                   & 81.2          & 83.4          & 78.5          & 74.8          & 78.3          & 68.4          & 74.9          & 76.2          & 77.0                 \\
AMLRIS~\citep{amlris}         & Swin-B\,/\,BERT-B             & 77.2          & 79.5          & 73.6          & 68.6          & 73.8          & 61.6          & 68.8          & 70.0          & 71.9                 \\
\rowcolor[HTML]{ECECEC}
\textbf{DeCo (Ours)}              & BEiT3-B$^*$\,/\,BEiT3-B$^*$           & \textbf{82.4} & \textbf{84.5} & \textbf{80.0} & \textbf{77.1} & \textbf{81.0} & \textbf{71.5} & \textbf{77.7} & \textbf{78.5} & \textbf{79.1}        \\
\midrule

PolyFormer~\citep{polyformer}         & Swin-L\,/\,BERT-B             & 76.0          & 78.3          & 73.3          & 69.3          & 74.6          & 61.9          & 69.2          & 70.2          & 71.6                 \\
UNINEXT~\citep{uninext}          & ConvNeXt-L\,/\,BERT-B           & 80.3          & 82.6          & 77.8          & 70.0          & 74.9          & 62.6          & 73.4          & 73.7          & 74.4                 \\
BarLeRIa~\citep{barleria} & CLIP-L$^*$\,/\,CLIP-L$^*$   & 79.0    & 80.8   & 77.0          & 74.2   & 77.8     & 68.3       & 72.7     & 73.3      & 75.4          \\
OneRef~\citep{oneref}        & BEiT3-L\,/\,BEiT3-L                   & 82.0          & 83.6          & 79.6          & 74.6          & 79.5          & 70.7          & 76.5          & 79.2          & 78.2                 \\
CMIRNet~\citep{cmirnet}       & Swin-L\,/\,BERT-B             & 78.2          & 80.4          & 75.2          & 69.8          & 75.3          & 62.1          & 70.4          & 72.1          & 72.9                 \\
DETRIS~\citep{detris}       & DINOv2-L$^*$\,/\,CLIP-B$^*$           & 80.7          & 82.2          & 77.9          & 73.1          & 77.5          & 66.0          & 73.2          & 74.7          & 75.7                 \\
InstanceVG~\citep{instancevg}     & BEiT3-L\,/\,BEiT3-L                   & \textbf{85.8} & \underline{86.7} & \textbf{84.1} & \underline{80.4}          & \underline{82.9}          & \underline{75.0}          & \underline{79.9}          & \underline{81.4}          & \underline{82.0}     \\
\rowcolor[HTML]{ECECEC}
\textbf{DeCo (Ours)}     & BEiT3-L$^*$\,/\,BEiT3-L$^*$   & \underline{85.4}          & \textbf{86.7}          & \underline{83.5}          & \textbf{81.9} & \textbf{85.0} & \textbf{76.9} & \textbf{81.7} & \textbf{82.6} & \textbf{83.0}     \\
\bottomrule

\end{tabular}}
\vspace{-10pt}
\end{table*}

%% file: Tables/3_other.tex
\begin{figure*}[t]
\centering

\vspace{-10pt}
\begin{minipage}[t]{0.51\textwidth}
\centering
    \footnotesize
    \captionof{table}{Main results of \textbf{REC} models using \textbf{P@.5} on G-Ref, ReferIt, and Flickr datasets.}
    \label{tab_other_rec}
    \vspace{-0.5em}
    \resizebox{1.0\columnwidth}{!}{
    \begin{tabular}{l|c|c|c|c}
    \toprule
    \multirow{2}{*}{Models} & Vision / Language & G-Ref   & ReferIt  & Flickr \\ \cmidrule{3-5}
        & Backbone         & val-g         & test          & test      \\
    \midrule
    SimVG~\citep{simvg}                   & BEiT3-B\,/\,BEiT3-B          & 78.8          & 74.8          & 82.0      \\
    OneRef~\citep{oneref}                  & BEiT3-B\,/\,BEiT3-B          & -             & \underline{77.2}     & \underline{83.6} \\
    MAP~\citep{map}                     & DINOv2-B$^*$\,/\,BERT-B$^*$  & 76.7          & 75.0          & 81.3      \\
    SwimVG~\citep{swimvg}                  & DINOv2-B$^*$\,/\,CLIP-B$^*$  & 79.1          & -             & 83.1      \\
    OneSVG~\citep{onesvg}                  & BEiT3-B$^*$\,/\,BEiT3-B$^*$          & \underline{81.5}          & -         & -     \\
    BARE~\citep{bare}                  & BEiT3-B$^*$\,/\,BEiT3-B$^*$          & -          & 75.3         & 82.3     \\
    \rowcolor[HTML]{ECECEC}
    \textbf{DeCo (Ours)}              & BEiT3-B$^*$\,/\,BEiT3-B$^*$          & \textbf{85.4} & \textbf{80.5} &  \textbf{84.4}    \\
    \midrule
    HiVG~\citep{hivg}                    & CLIP-L$^*$\,/\,CLIP-L$^*$           & -             & 76.2     & 82.2 \\
    SimVG~\citep{simvg}                   & BEiT3-L\,/\,BEiT3-L          & 80.4          & 78.8          & 83.2      \\
    OneRef~\citep{oneref}                  & BEiT3-L\,/\,BEiT3-L          & -             & \underline{81.1}     & \underline{84.8} \\
    OneSVG~\citep{onesvg}                  & BEiT3-L$^*$\,/\,BEiT3-L$^*$          & \underline{87.2}          & -         & -    \\
    BARE~\citep{bare}                  & BEiT3-L$^*$\,/\,BEiT3-L$^*$          & -          & 80.0         & 83.3     \\
    \rowcolor[HTML]{ECECEC}
    \textbf{DeCo (Ours)}              & BEiT3-L$^*$\,/\,BEiT3-L$^*$          & \textbf{90.3} & \textbf{83.7} &  \textbf{86.7}         \\
    \bottomrule
    \end{tabular}}
    % \vspace{-15pt}    
\end{minipage}
\hspace{0.5mm}
\begin{minipage}[t]{0.47\textwidth}
\centering
    \footnotesize
    \captionof{table}{Main results of \textbf{RES} models using \textbf{oIoU} and \textbf{mIoU} on the G-Ref dataset.}
    \label{tab_other_res}
    \vspace{-0.5em}
    \setlength{\tabcolsep}{6.5pt}
    \resizebox{1.0\columnwidth}{!}{
    \begin{tabular}{l|c|cc}
    \toprule
    \multirow{2}{*}{Models} & Vision / Language & \multicolumn{2}{c}{G-Ref (val-g)} \\
    \cmidrule{3-4}
    & Backbone         & oIoU            & mIoU            \\
    \midrule
    BarLeRIa~\citep{barleria}  & CLIP-B$^*$\,/\,CLIP-B$^*$    & 61.6            & -               \\
    LQMFormer~\citep{lqmformer}  & Swin-B\,/\,BERT-B    & 63.0            & -               \\
    Prompt-RIS~\citep{promptris} & SAM-B$^*$\,/\,CLIP-B$^*$     & \underline{64.8}   & \underline{69.2}            \\
    ASDA~\citep{asda}                    & CLIP-B\,/\,CLIP-B    & 63.6            & -               \\
    CMANet~\citep{cmanet}                  & Swin-B\,/\,BERT-B    & 64.1            & -               \\
    DETRIS~\citep{detris}                  & DINOv2-B$^*$\,/\,CLIP-B$^*$  & 63.2            & 65.9            \\
    MiCA~\citep{mica}                    & DINOv2-B$^*$\,/\,CLIP-B$^*$  & -               & 66.7            \\
    WIMFRIS~\citep{wimfris}                 & DINOv2-B$^*$\,/\,CLIP-B$^*$  & -               & 67.1            \\
    \rowcolor[HTML]{ECECEC}
    \textbf{DeCo (Ours)}     & BEiT3-B$^*$\,/\,BEiT3-B$^*$          & \textbf{72.3}   & \textbf{73.7}   \\
    \midrule
    DETRIS~\citep{detris}                  & DINOv2-L$^*$\,/\,CLIP-B$^*$  & \underline{65.0}            & 67.9            \\
    MiCA~\citep{mica}                    & DINOv2-L$^*$\,/\,CLIP-B$^*$  & -               & 68.6            \\
    WIMFRIS~\citep{wimfris}                 & DINOv2-L$^*$\,/\,CLIP-B$^*$  & -               & \underline{69.3}            \\
    \rowcolor[HTML]{ECECEC}
    \textbf{DeCo (Ours)}     & BEiT3-L$^*$\,/\,BEiT3-L$^*$          & \textbf{78.4}   & \textbf{79.1}  \\
    \bottomrule
    \end{tabular}}
    % \vspace{-15pt}	
\end{minipage}

\vspace{2.5mm}

\begin{minipage}[t]{0.49\textwidth}
\centering
    \footnotesize
    \captionof{table}{Ablation on core components of DeCo.}
    \label{tab_abla_comp}
    \vspace{-0.5em}
    \setlength{\tabcolsep}{5pt}
    \resizebox{1.0\columnwidth}{!}{
    \begin{tabular}{l|rr|cc|cc|c}
    \toprule
    \multirow{2}{*}{\diagbox[width=7.2em,height=2.8em]{Models}{\shortstack{test-u}}} 
    & \multicolumn{2}{c|}{Trainable Head} & \multicolumn{2}{c|}{REC} & \multicolumn{2}{c|}{RES} & \multirow{2}{*}{Avg.} \\ \cmidrule{2-7}
    & Params.  & FLOPs  & P@.5  & P@.5:.95  & oIoU   & mIoU  &     \\
    \midrule
    Previous SOTA  & -     & -     & 83.2     & -  & 66.8 & 69.7     & -  \\
    \midrule
    LoM   & 7.9M  & 15.9G & 85.5 & 67.2 & 71.2 & 72.0 & 74.0   \\
    SoM   & \textbf{3.4M}  & \textbf{13.3G}  & 84.8 & 66.7 & 71.6 & 72.5 & 73.9   \\
    DoM   & 8.3M  & 17.6G & 85.6 & 67.4 & 71.8 & 72.6 & 74.4   \\
    DoM \textit{w/} TSD  & 10.6M & 24.2G & 86.3 & 69.1 & \underline{72.5} & \underline{73.6} & 75.4  \\
    DoM \textit{w/} HPC  & 9.3M  & 17.6G & \underline{86.5} & \underline{70.5} & 72.2 & 73.3 & \underline{75.6}  \\
    \rowcolor[HTML]{ECECEC}
    \textbf{DeCo}  & 11.6M & 24.2G & \textbf{87.0} & \textbf{71.7} & \textbf{73.0} & \textbf{74.5} & \textbf{76.6}  \\
    \bottomrule
    \end{tabular}}
    % \vspace{-15pt}    
\end{minipage}
\hspace{0.5mm}
\begin{minipage}[t]{0.49\textwidth}
\centering
    \footnotesize
    \captionof{table}{Ablation on PEFT and LoRA $r$ settings.}
    \label{tab_abla_peft}
    \vspace{-0.5em}
    \setlength{\tabcolsep}{4.1pt}
    \resizebox{1.0\columnwidth}{!}{
    \begin{tabular}{l|rr|cc|cc|c}
    \toprule
    \multirow{2}{*}{\diagbox[width=7.2em,height=2.8em]{Methods}{\shortstack{test-u}}}
    & \multicolumn{2}{c|}{Trainable Backbone} & \multicolumn{2}{c|}{REC} & \multicolumn{2}{c|}{RES} & \multirow{2}{*}{Avg.} \\ \cmidrule{2-7}
    & Params.  & FLOPs  & P@.5  & P@.5:.95  & oIoU   & mIoU  &   \\ 
    \midrule
    Full Tuning\hspace{1.2em}    & 221.3M & 81.0G & 72.6 & 53.8 & 57.8 & 61.2 & 61.4 \\
    Frozen      & 0.0M   & 0.0G  & 68.1 & 43.5 & 51.8 & 53.0 & 54.1 \\
    \midrule
    LoRA ($r=4$)  & \textbf{0.6M}   & \textbf{0.3G}  & 86.3 & 69.0 & 72.0 & 72.8 & 75.0  \\
    LoRA ($r=8$)  & 1.2M   & 0.5G  & 86.5 & 69.7 & 72.2 & 73.2 & 75.4   \\
    LoRA ($r=16$) & 2.4M   & 1.0G  & 86.6 & 70.3 & 72.7 & 73.9 & 75.9  \\
    \rowcolor[HTML]{ECECEC}
    \textbf{LoRA ($r=\textbf{32}$)} & 4.7M   & 1.9G  & \underline{87.0} & \underline{71.7}  & \textbf{73.0}  & \textbf{74.5}  & \textbf{76.6} \\
    LoRA ($r=64$) & 9.4M   & 3.8G  & \textbf{87.1} & \textbf{71.8} & \underline{73.0} & \underline{74.4} & \underline{76.6}   \\ 
    \bottomrule
    \end{tabular}
    }
    % \vspace{-15pt}	
\end{minipage}

\vspace{2.5mm}

\begin{minipage}[t]{0.49\textwidth}
\centering
    \footnotesize
    \captionof{table}{Ablation on LoRA inserted positions.}
    \label{tab_abla_pos}
    \vspace{-0.5em}
    \setlength{\tabcolsep}{4.2pt}
    \resizebox{1.0\columnwidth}{!}{
    \begin{tabular}{l|rr|cc|cc|c}
    \toprule
    \multirow{2}{*}{\diagbox[width=7em,height=2.8em]{Positions}{\shortstack{test-u}}}
    & \multicolumn{2}{c|}{Trainable Backbone} & \multicolumn{2}{c|}{REC} & \multicolumn{2}{c|}{RES} & \multirow{2}{*}{Avg.} \\ \cmidrule{2-7}
    & Params.  & FLOPs  & P@.5  & P@.5:.95  & oIoU   & mIoU  &    \\
    \midrule
    Q                & 1.2M  & 0.5G & 84.3 & 65.9 & 70.4 & 71.0 & 72.9 \\
    V                & \textbf{1.2M}  & \textbf{0.5G} & 86.1 & 69.0 & 71.9 & 72.8 & 75.0 \\
    Q-V              & 2.4M  & 1.0G & 86.7 & 70.1 & 72.6 & 73.9 & 75.8 \\
    \rowcolor[HTML]{ECECEC}
    \textbf{Q-K-V-O} & 4.7M  & 1.9G & \underline{87.0} & \underline{71.7} & \underline{73.0} & \underline{74.5} & \underline{76.6} \\
    FFN              & 5.9M  & 2.4G & 86.5 & 70.6 & 72.7 & 74.2 & 76.0 \\
    Q-K-V-O-FFN      & 10.6M & 4.3G & \textbf{87.1} & \textbf{71.8} & \textbf{73.2} & \textbf{74.8} & \textbf{76.7} \\ \bottomrule
    \end{tabular}
    }
    % \vspace{-15pt}	
\end{minipage}
\hspace{0.5mm}
\begin{minipage}[t]{0.49\textwidth}
\centering
    \footnotesize
    \captionof{table}{Ablation on various segmentation heads.}
    \label{tab_abla_seg}
    \vspace{-0.5em}
    \setlength{\tabcolsep}{5.8pt}
    \resizebox{1.0\columnwidth}{!}{
        \begin{tabular}{l|rr|cc|cc|c}
    \toprule
    \multirow{2}{*}{\diagbox[width=5.8em,height=2.9em]{Models}{\shortstack{test-u}}} 
    & \multicolumn{2}{c|}{Trainable Head} & \multicolumn{2}{c|}{REC} & \multicolumn{2}{c|}{RES} & \multirow{2}{*}{Avg.} \\ \cmidrule{2-7}
    & Params.  & FLOPs  & P@.5  & P@.5:.95  & oIoU   & mIoU  &    \\
    \midrule
    \rowcolor[HTML]{ECECEC}
    \textbf{UNet}      & \textbf{0.4M}  & \textbf{2.6G}  & \textbf{87.0} & \textbf{71.7} & 73.0 & 74.5 & \textbf{76.6} \\
    SegFormer & 0.5M  & 5.0G  & 86.5 & 70.9 & 72.9 & 74.1 & 76.1 \\
    UPerNet   & 7.5M  & 41.6G & \underline{86.7} & 71.2 & \underline{73.3} & \underline{74.6} & 76.5 \\
    ReMamber  & 15.0M & 62.1G & 86.7 & 71.0 & 72.7 & 74.4 & 76.2 \\
    LAVT-RS   & 5.3M  & 29.7G & 86.6 & \underline{71.3} & \textbf{73.3} & \textbf{74.7} & \underline{76.5} \\
    DETRIS    & 23.9M & 37.2G & 86.3 & 71.1 & 72.7 & 74.2 & 76.1 \\
    \bottomrule
    \end{tabular}
    }
    % \vspace{-15pt}	
\end{minipage}

\vspace{1.5mm}

\includegraphics[width=0.98\linewidth]{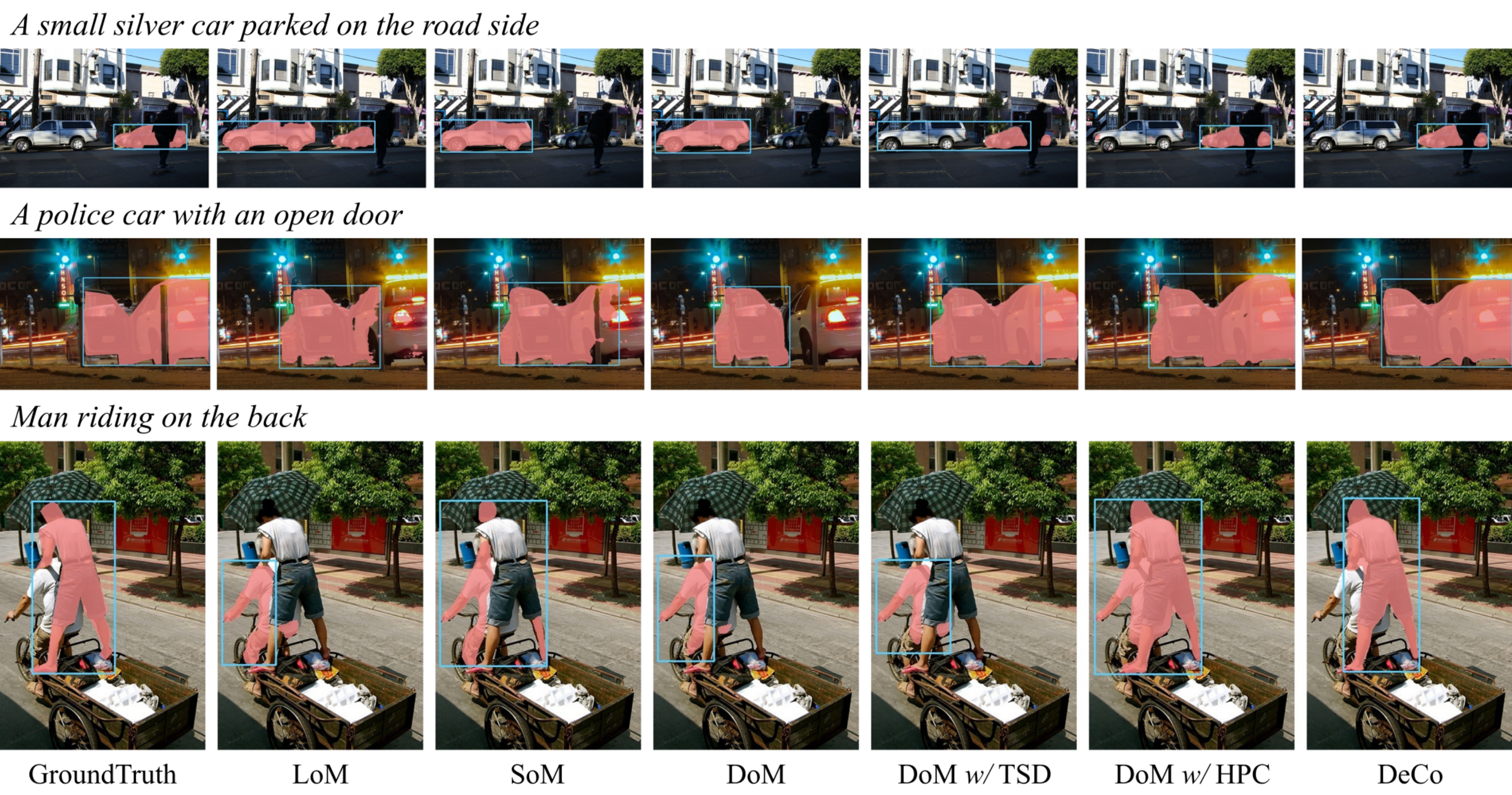}
\vspace{-5pt}
\captionof{figure}{Qualitative results on core components of DeCo on G-Ref (test-u).}
\label{fig_comp}

\vspace{-11pt}
\end{figure*}

%% file: Sections/5_conclusion.tex
\vspace{-6pt}
\section{Conclusion}
\vspace{-4pt}

We presented DeCo, an efficient decouple-to-couple learning framework for unified referring expression comprehension and segmentation. By first decoupling shared visual representations into task-specific features and then coupling complementary semantic and spatial priors, DeCo enables each task to exploit its preferred visual cues while benefiting from cross-task knowledge. Extensive experiments demonstrate consistent improvements over existing methods across multiple referring expression benchmarks and settings, with only lightweight adaptation of the pretrained backbone. Our results suggest that explicitly managing the tension between task-specificity and task complementarity provides an effective and efficient paradigm for multi-task visual grounding.

%% file: Sections/6_appendix.tex
\section*{Appendix}

\section{Details of Efficient Multimodal Encoding}
\label{sec_app_encoding}

As introduced in \cref{sec_overview}, DeCo employs a frozen BEiT3~\citep{beit3} encoder with the classic Low-Rank Adaptation (LoRA)~\citep{lora} followed by the Simple Feature Pyramid (SimFPN)~\citep{vitdet} to construct efficient multimodal and multi-scale representations. Here, we provide more architectural details of these two components.

\paragraph{Low-Rank Adaptation.} 
To efficiently exploit the pretrained multimodal encoder while preserving its original knowledge, we keep all BEiT3 parameters frozen and inject LoRA into the linear projections of its Transformer layers. Formally, at the $i$-th layer, the visual patch embeddings $\mathbf{V}_i$ and linguistic token embeddings $\mathbf{T}_i$ are first concatenated into a unified multimodal representation $\mathbf{Z}_i$, which is then processed by the multi-head self-attention operator. Given a fixed projection matrix $\mathbf{W}_0\in\mathbb{R}^{d\times d}$, LoRA models its task-specific update with a pair of low-rank projections:
\begin{equation}
\mathbf{W}
=
\mathbf{W}_0+\Delta\mathbf{W},
\qquad
\Delta\mathbf{W}
=
\frac{\alpha}{r}\mathbf{C}\mathbf{A},
\end{equation}
where $\mathbf{A}\in\mathbb{R}^{r\times d}$ and $\mathbf{C}\in\mathbb{R}^{d\times r}$ are two lightweight trainable matrices, $r\ll d$ denotes the low-rank dimension, and the scaling factor $\alpha$ controls the update ratio. Accordingly, the adapted projection of the multimodal tokens $\mathbf{Z}_i$ is formulated as follows:
\begin{equation}
\mathbf{Z}_i
=
\mathbf{Z}_i\mathbf{W}_0^{\top} + 
\frac{\alpha}{r} \mathbf{Z}_i\,\mathbf{A}^{\top}\mathbf{C}^{\top}.
\end{equation}
In DeCo, LoRA is inserted into the query, key, value, and output projections of each Transformer attention layer. For BEiT3-B and BEiT3-L, the embedding dimension $d$ is $768$ and $1,024$, with $12$ and $24$ layers, respectively. Unless otherwise specified, we set both $r$ and $\alpha$ to $32$ for all experiments. Since these projections operate on the unified vision-language sequence, the low-rank updates efficiently enable multimodal interactions while avoiding retraining the original encoder.

\paragraph{Simple Feature Pyramid.} 
The final visual representation produced by BEiT3 is a single-scale sequence of patch embeddings. To provide the multi-scale representations required by localization and segmentation, we adopt SimFPN to construct a four-level feature pyramid. Specifically, excluding the class token $\bar{\mathbf{v}}$, we first reshape the visual patch embeddings into a spatial feature map:
\begin{equation}
\mathbf{F}
=
\operatorname{reshape}
\big(
[\mathbf{v}_1,\ldots,\mathbf{v}_n]
\big)
\in
\mathbb{R}^{\frac{h}{p_{size}}\times\frac{w}{p_{size}}\times d},
\end{equation}
where $p_{size} = 16$ represents the patch size. Then, $\mathbf{F}$ is processed by four parallel deconvolutions and convolutions to obtain features at different spatial resolutions, as follows:
\begin{equation}
\mathbf{V}_l
=
\phi_l
\big(
\mathcal{S}_l(\mathbf{F})
\big)
\in
\mathbb{R}^{h_l\times w_l\times c},
\qquad l\in\{1,2,3,4\},
\end{equation}
where $\mathcal{S}_l(\cdot)$ denotes the corresponding rescaling operation, $\phi_l(\cdot)$ projects each feature map into a small channel dimension $c=256$, and the four scale factors $s_l \in \{4, 2, 1, \frac{1}{2}\}$ yield different spatial resolutions $(h_l,w_l) =s_l\,(\frac{h}{p_{size}},\frac{w}{p_{size}})$, resulting in multi-scale features $\mathcal{V} = \{\mathbf{V}_l\}_{l=1}^4$. This design bridges the single-scale representation of the frozen multimodal encoder and the multi-resolution requirements of localization and segmentation, while its lightweight modules avoid a costly hierarchical backbone and balance spatial representation capability with computational efficiency.

\section{More Experimental Setup Details}

\input{Tables/append_dst_details}

\subsection{Datasets}
\label{sec_app_dataset}

To investigate the effectiveness of DeCo, we conduct extensive experiments across diverse vision-language benchmarks covering natural and remote sensing images. The datasets differ in visual appearance, linguistic complexity, image modality, and task formulation, providing a comprehensive testbed for evaluating multi-task visual grounding. Detailed statistics are summarized in \cref{tab_dst_details}.

\subsubsection{Natural Image Benchmarks} 

We first evaluate DeCo on five widely used natural image benchmarks: \textbf{RefCOCO}~\citep{refcoco} is derived from MSCOCO~\citep{mscoco} and contains relatively concise referring expressions for identifying objects in diverse everyday scenes. \textbf{RefCOCO+}~\citep{refcoco} follows the same image source but removes location-based expressions, placing greater emphasis on appearance-related attributes and therefore requiring finer-grained correspondence. \textbf{G-Ref}~\citep{gref1, gref2} provides substantially longer and more descriptive expressions, with an average of 8.4 words per expression, making it a challenging benchmark for multi-task visual grounding. It provides two different partitions, UMD~\citep{gref2} and Google~\citep{gref1}, with the former adopted as the main training and evaluation protocol in our experiments. \textbf{ReferIt}~\citep{referit} contains referring expressions collected for objects and regions in natural images and provides a large test set for evaluating generalization beyond MSCOCO-based benchmarks. \textbf{Flickr}~\citep{flickr30k} constructs large-scale image-phrase pairs with relatively short descriptions, providing an additional benchmark for evaluating REC in diverse natural scenes. 

Following previous works~\citep{oneref, propvg, amlris}, we conduct experiments on both the \textit{Standard} and \textit{Mixed} settings. In the standard setting, each dataset is trained and evaluated independently using its corresponding splits. In the mixed setting, we combine the training sets of RefCOCO, RefCOCO+, and the UMD partition of G-Ref to construct a unified training set, while keeping their original validation and test splits separate for evaluation.

\subsubsection{Remote Sensing Image Benchmarks} 

To further examine the generalization of DeCo beyond natural images, we consider five remote sensing benchmarks, including optical, SAR, and UAV imagery. \textbf{DIOR-RSVG}~\citep{mgvlf} is an optical visual grounding benchmark, where referring expressions are used to identify specific objects in aerial scenes. Compared with natural images, the large image scale, dense object distributions, and substantial appearance variations introduce additional challenges for localization. \textbf{SARVG1.0}~\citep{tacmt} focuses on visual grounding in SAR imagery. In contrast to optical imagery, SAR observations exhibit distinct scattering characteristics and substantially different visual patterns, providing a challenging testbed for evaluating the robustness of DeCo under various imaging modalities. \textbf{RRSIS-D}~\citep{rmsin} is an optical referring image segmentation benchmark that associates textual description with pixel-level object masks. It therefore evaluates whether language-guided understanding can be translated from object-level localization to fine-grained spatial segmentation in aerial scenes. \textbf{RIS-LAD}~\citep{ris_lad} further considers UAV imagery and contains substantially longer referring expressions, with an average length of 19.6 words. Its fine-grained descriptions and aerial perspectives provide a more challenging setting for evaluating tiny object segmentation. \textbf{RefDIOR}~\citep{rrsecs} provides a unified benchmark covering both object localization and pixel-level segmentation in optical remote sensing imagery. We use it to evaluate the unified referring remote sensing comprehension and segmentation capability of DeCo, thereby extending the joint REC and RES formulation to remote sensing scenarios. All the above remote sensing image benchmarks follow the standard setting for training, validation, and testing.

\subsection{Evaluation Metrics}
\label{sec_app_metric}

In the main experiments, we evaluate DeCo using commonly used metrics for REC, RES, and their corresponding remote sensing variants, RSVG and RRSIS. For REC, we adopt Precision@IoU threshold of 0.5 (\textbf{P@.5}), which measures the percentage of predictions whose IoU with the ground-truth box exceeds 0.5. Specifically, given the $i$-th predicted bounding box $\mathbf{B}_i$ and its corresponding ground-truth box $\mathbf{B}_i^*$, P@.5 can be calculated as follows:
\begin{equation}
\operatorname{IoU}(\mathbf{B}_i,\mathbf{B}_i^*)
=
\frac{|\mathbf{B}_i\cap\mathbf{B}_i^*|}
{|\mathbf{B}_i\cup\mathbf{B}_i^*|}, \qquad
{\rm P\text{@.5}}
=
\frac{1}{N}
\sum_{i=1}^{N}
\mathbb{I}
\big [
\operatorname{IoU}
(\mathbf{B}_i,\mathbf{B}_i^*)
\geq \text{0.5}
\big ],
\end{equation}
where $N$ denotes the number of referring expressions and $\mathbb{I}[\cdot]$
is the indicator function. For RES, we report overall IoU (\textbf{oIoU}) and mean IoU (\textbf{mIoU}). Given the predicted mask $\mathbf{M}_i$ and the ground-truth mask $\mathbf{M}_i^*$, oIoU aggregates the intersection and union across all samples, while mIoU averages the per-sample IoU, which can be formulated as follows:
\begin{equation}
\operatorname{oIoU}
=
\frac{
\sum_{i=1}^{N}|\mathbf{M}_i\cap\mathbf{M}_i^*|
}{
\sum_{i=1}^{N}|\mathbf{M}_i\cup\mathbf{M}_i^*|
},
\qquad
\operatorname{mIoU}
=
\frac{1}{N}
\sum_{i=1}^{N}
\frac{
|\mathbf{M}_i\cap\mathbf{M}_i^*|
}{
|\mathbf{M}_i\cup\mathbf{M}_i^*|
}.
\end{equation}

For a more precise evaluation in the ablation studies, we additionally
introduce \textbf{P@.5:.95} for REC, which averages the precision over IoU thresholds from 0.5 to 0.95 with an interval of 0.05. The metrics used for RSVG and RRSIS follow similar computation protocols described above. All metrics are reported in percentage form, with higher values indicating better performance.

\input{Tables/append_imp_details}
\subsection{Implementation Details}
\label{sec_app_implem}

We instantiate DeCo with two model scales, termed \textbf{DeCo-B} and \textbf{DeCo-L}, using frozen BEiT3-B/L~\citep{beit3} with LoRA~\citep{lora} adaptation as the multimodal encoders, respectively. For localization, we adopt a decoder-only Deformable DETR~\citep{deformabledetr} consisting of four Transformer decoder layers, with a feature dimension of $256$ and an FFN hidden dimension of $1,024$. For segmentation, we employ a lightweight UNet~\citep{unet} with an input feature dimension of $256$ and a hidden embedding dimension of $32$. For datasets without mask annotations, including ReferIt, Flickr, DIOR-RSVG, and SARVG1.0, we use box-filled binary masks derived from the ground-truth bounding boxes as pseudo-mask supervision. We use Adam~\citep{adam} as the optimizer and MultiStepLR as the learning rate scheduler for all benchmarks. The dataset-specific training configurations are summarized in \cref{tab_implem_details}. The maximum text length is set to $20$ tokens for all benchmarks except RIS-LAD, for which we use $40$ tokens due to its substantially longer referring expressions. All experiments are conducted on eight NVIDIA RTX 4090 GPUs. Different batch sizes and learning rates are adopted for DeCo-B and DeCo-L according to their model scales and the input resolutions of various benchmarks.

\section{More Experimental Results}
\label{sec_app_exps}

\input{Tables/append_res_miou}

\subsection{Main Results for Referring Expression Segmentation using mIoU}

\begin{figure*}[t]
    \centering
    % \vspace{-10pt}
    \includegraphics[width=1.0\linewidth]{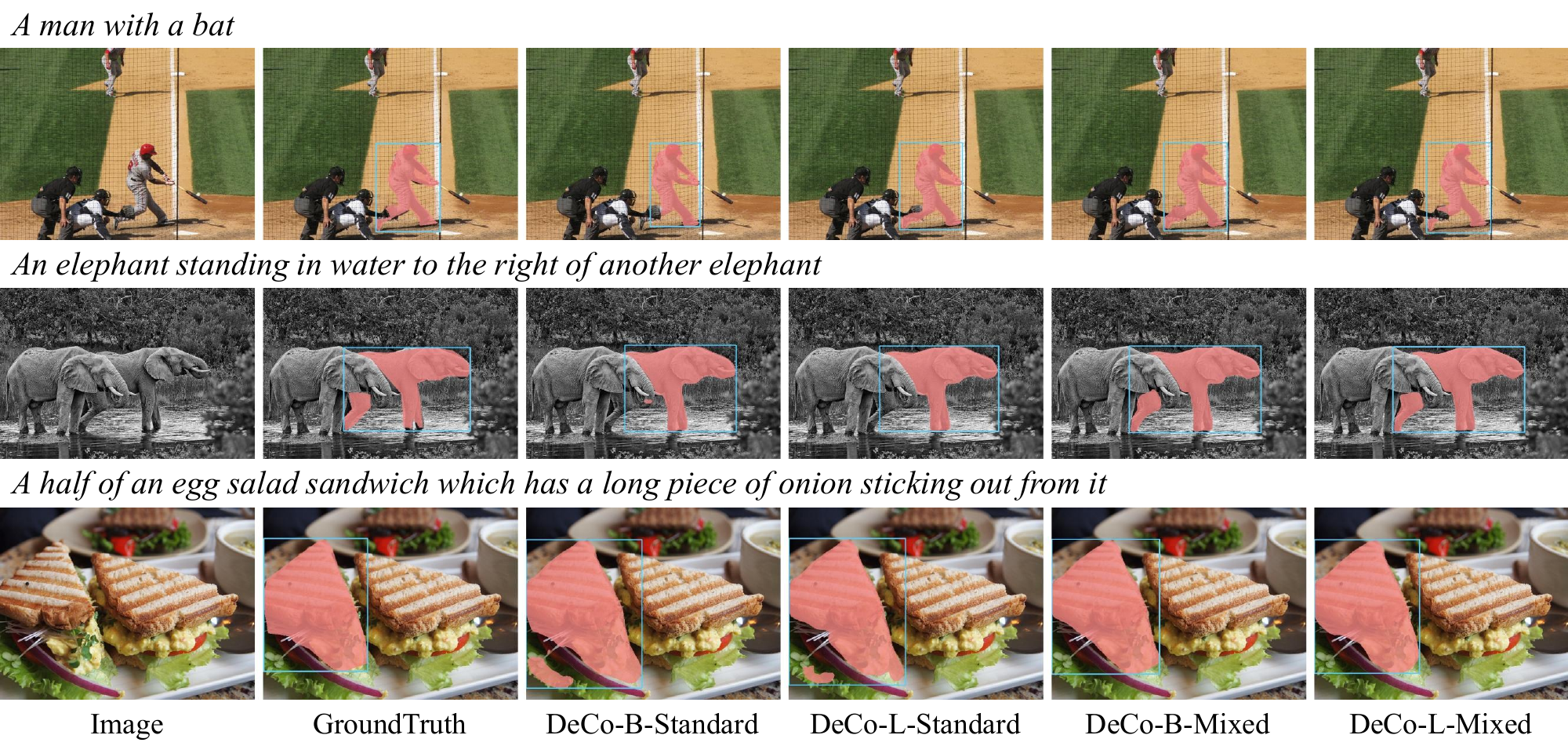}
    \caption{Qualitative results of the proposed DeCo across different settings on G-Ref (test-u).}
    \label{fig_main_qr_testu}
    % \vspace{-10pt}
\end{figure*}

\cref{tab_res_miou} reports the RES results using mIoU on RefCOCO, RefCOCO+, and G-Ref. Overall, DeCo remains highly competitive under this sample-wise segmentation metric, showing a consistent trend with the oIoU results reported in the main paper. Under the standard setting, DeCo-B achieves an average mIoU of 76.7\%, outperforming the strongest competing method, PropVG~\citep{propvg}, by 2.6\%. Scaling to DeCo-L further increases the average mIoU to 80.7\%, yielding a substantial 4.3\% gain over OneRef~\citep{oneref}. Notably, DeCo-L consistently ranks first across all eight evaluation splits, demonstrating that the proposed decouple-to-couple learning effectively improves segmentation quality across samples rather than benefiting only from large foreground regions. 

Under the mixed setting, DeCo-B obtains an average mIoU of 80.1\%, surpassing the previous best result of SaFiRe~\citep{safire} by 1.1\% and achieving the best performance on all evaluation splits. DeCo-L further reaches an average mIoU of 83.6\%. Although InstanceVG~\citep{instancevg} performs better on the RefCOCO splits, DeCo-L consistently delivers superior results on the more challenging RefCOCO+ and G-Ref datasets, leading to the best overall average performance. The advantage also extends to the G-Ref Google split, where DeCo-B and DeCo-L achieve 73.7\% and 79.1\% mIoU, exceeding WIMFRIS~\citep{wimfris} by 6.6\% and 9.8\%, as listed in \cref{tab_other_res}. 

Moreover, \cref{fig_main_qr_testu} provides qualitative comparisons across various model scales and training settings, showing that DeCo consistently produces well-aligned localization boxes and accurate masks for both simple and challenging referring expressions. These results confirm the effectiveness and robustness of DeCo under both dataset-specific and mixed-data training settings.

\subsection{Generalization to Remote Sensing Scenarios}

Beyond the standard natural image benchmarks, we further investigate the generalization of DeCo to diverse remote sensing scenarios. Compared with natural images, remote sensing imagery exhibits substantial domain shifts caused by aerial viewpoints, large-scale variations, dense object distributions, and modality-specific visual characteristics, posing additional challenges to multi-task visual grounding. In this domain, REC and RES are extended to remote sensing visual grounding (RSVG)~\citep{rsvg} and referring remote sensing image segmentation (RRSIS)~\citep{lgce}, respectively. To provide a comprehensive evaluation, we consider multiple benchmarks spanning optical, SAR, and UAV imagery, enabling us to assess whether the proposed method remains effective under substantially different visual distributions and imaging modalities.

\input{Tables/append_rsvg}

\textbf{Main results for RSVG on DIOR-RSVG.} 
\cref{tab_dior_rsvg} reports the results on the DIOR-RSVG test set. DeCo-B already achieves competitive performance, obtaining 85.2\% P@.5 and 76.0\% mIoU, which are on par with the strongest existing methods. Scaling to DeCo-L further yields consistent improvements across all evaluation metrics. Specifically, DeCo-L achieves 86.9\%, 84.3\%, 79.0\%, 69.2\%, and 47.6\% under P@.5--P@.9, respectively, while reaching 86.2\% oIoU and 78.6\% mIoU. Compared with the recent advanced method, VisproVG~\citep{visprovg}, DeCo-L improves P@.5, P@.7, P@.9, oIoU, and mIoU by 1.7\%, 0.8\%, 0.7\%, 1.5\%, and 2.6\%, respectively. The consistent gains across increasingly strict IoU thresholds indicate that DeCo not only identifies the referred remote sensing objects reliably but also produces more precise localization.

\textbf{Main results for RSVG on SARVG1.0.} 
\cref{tab_sarvg1.0} lists the results on the SARVG1.0 test set, which further examines the generalization of DeCo from optical imagery to SAR modality. Despite the distinct scattering characteristics and visual appearance of SAR, both DeCo-B and DeCo-L achieve strong performance across all evaluation metrics. DeCo-B reaches 89.6\% P@.5 and 83.8\% mIoU, already surpassing the previous best method, TACMT~\citep{tacmt}, on all reported precision thresholds. DeCo-L further improves the performance to 91.2\% P@.5 and 86.3\% mIoU. The advantage becomes more evident under stricter localization criteria. At P@.9, DeCo-B and DeCo-L achieve 73.5\% and 75.1\%, outperforming TACMT by 5.2\% and 6.8\%, respectively. These consistent gains across various IoU thresholds demonstrate that DeCo can effectively transfer its language-guided localization capability to SAR imagery while maintaining high localization precision.

\textbf{Main results for RSVG on RefDIOR.} 
\cref{tab_refdior_rsvg} summarizes the localization results on the RefDIOR test set. RefDIOR is designed for multi-task visual grounding in remote sensing scenes, making it a more comprehensive benchmark for evaluating our DeCo. DeCo-B achieves 83.1\% P@.5 and 75.0\% mIoU, surpassing the previous best method, CCFormer~\citep{rrsecs}, across all reported metrics. DeCo-L further improves the results to 85.3\% P@.5 and 77.0\% mIoU. The superiority of DeCo remains consistent under stricter IoU thresholds. Compared with CCFormer, DeCo-L improves P@.5, P@.7, and P@.9 by 2.9\%, 3.8\%, and 4.2\%, respectively, while gaining 1.8\% in oIoU and 2.9\% in mIoU. These results indicate that the benefits of modeling task-specificity and cross-task complementarity also extend to unified multi-task learning in remote sensing scenarios.

\input{Tables/append_rrsis}

\textbf{Main results for RRSIS on RRSIS-D.} 
Different from natural referring segmentation, RRSIS requires language-guided pixel-level understanding under overhead viewpoints and complex object layouts, making it a challenging testbed for evaluating DeCo in remote sensing segmentation. As shown in \cref{tab_rrsis_d}, DeCo-B achieves 77.3\% P@.5, 78.3\% oIoU, and 66.4\% mIoU, outperforming the previous best method, MCD-Net~\citep{mcdnet}, on the major evaluation metrics. DeCo-L further improves the performance to 81.3\% P@.5, 79.7\% oIoU, and 68.9\% mIoU. Compared with the strongest prior results~\citep{s2lcnet, mcdnet}, DeCo-L improves P@.5, P@.6, and P@.7 by 5.7\%, 5.7\%, and 3.8\%, respectively, while increasing oIoU and mIoU by 1.6\% and 3.8\%. These results show that the decouple-to-couple design transfers effectively to remote sensing referring segmentation and provides more accurate region-level alignment.

\textbf{Main results for RRSIS on RIS-LAD.} 
\cref{tab_ris_lad} provides the comparison on the RIS-LAD test set. RIS-LAD is a challenging UAV-based benchmark with substantially longer textual expressions and complex small objects, which impose greater demands on fine-grained language understanding and precise spatial localization. Despite these challenges, DeCo-B achieves 51.1\% P@.5, 50.3\% oIoU, and 47.1\% mIoU, outperforming the previous best method, SAARN~\citep{ris_lad}, by 6.1\%, 0.7\%, and 5.4\%, respectively. DeCo-L further improves these results to 56.1\% P@.5, 54.8\% oIoU, and 50.2\% mIoU, while retaining its advantage under stricter IoU thresholds. Such substantial gains demonstrate the ability of DeCo to preserve semantic discrimination and spatial precision in challenging UAV scenarios with long referring expressions and small targets.

\textbf{Main results for RRSIS on RefDIOR.} 
\cref{tab_refdior_rrsis} presents the segmentation results on the RefDIOR test set. As a unified multi-task remote sensing benchmark, RefDIOR provides a challenging setting for jointly evaluating localization and segmentation. DeCo-B achieves 81.6\% P@.5, 81.8\% oIoU, and 71.6\% mIoU, outperforming CCFormer~\citep{rrsecs} by 1.5\%, 0.9\%, and 0.6\%, respectively. DeCo-L further improves these results to 83.2\% P@.5, 83.8\% oIoU, and 73.4\% mIoU, bringing impressive gains of 3.1\%, 2.9\%, and 2.4\% over CCFormer, respectively. Together with the RSVG results on the same benchmark, all the improvements demonstrate that our decouple-to-couple paradigm is well suited to unified multi-task learning in remote sensing scenarios.

\begin{figure*}[t]
    \centering
    % \vspace{-10pt}
    \includegraphics[width=1.0\linewidth]{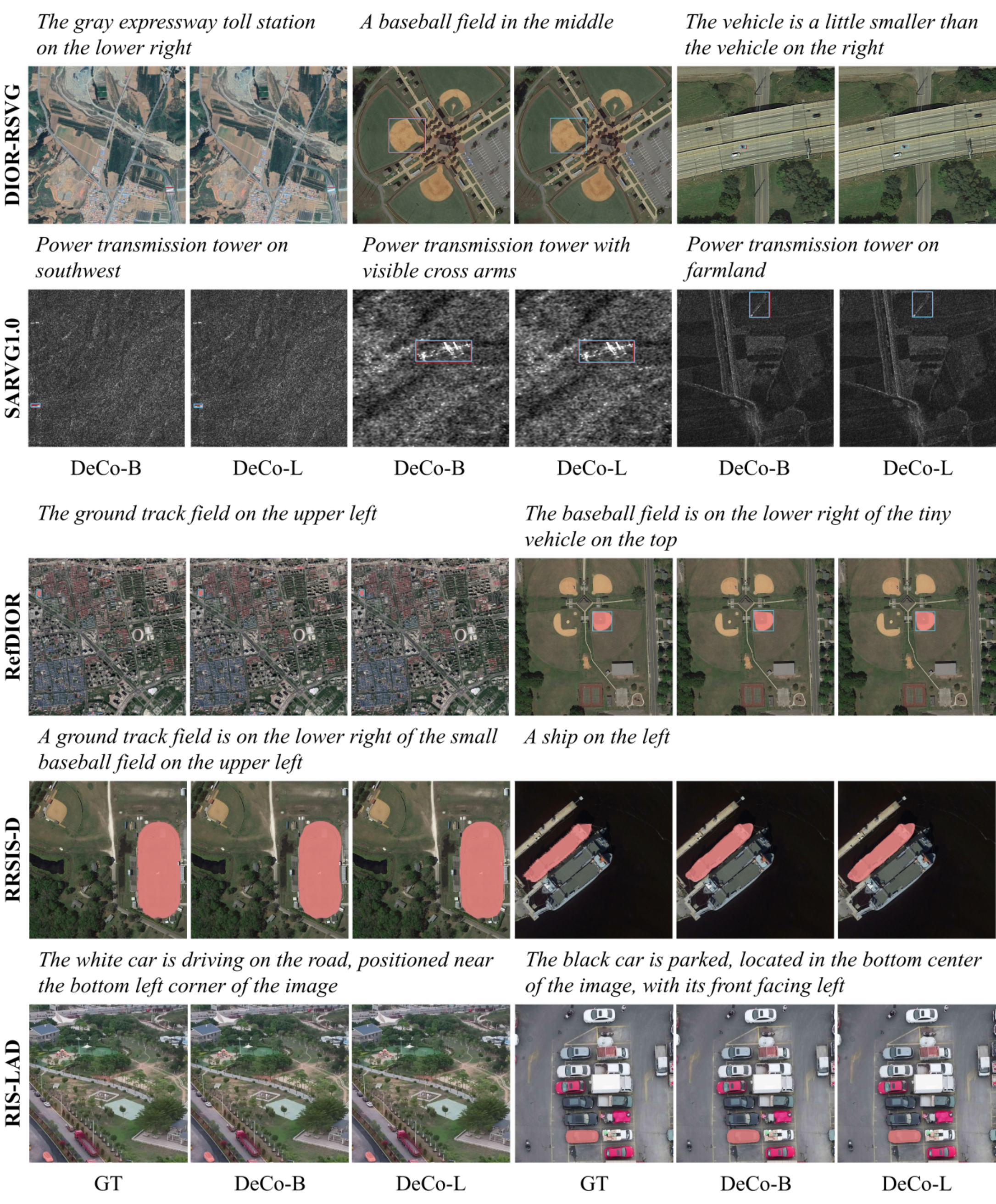}
    \caption{Qualitative results of our DeCo across different remote sensing image benchmarks.}
    \label{fig_main_rs_test}
    \vspace{-5pt}
\end{figure*}

\cref{fig_main_rs_test} shows qualitative results across diverse remote sensing benchmarks. DeCo reliably grounds the referred objects under optical, SAR, and UAV imagery, including small objects and relation-intensive expressions, while producing well-aligned localization boxes and segmentation masks. Overall, the above six quantitative results reveal that existing methods developed for natural images generally lag behind specialized remote sensing approaches, indicating a clear domain gap between the two scenarios. In contrast, DeCo consistently bridges this gap and achieves superior performance across various tasks and modalities. Moreover, existing remote sensing methods typically rely on full tuning, which introduces substantial optimization burden when processing high-resolution images. By adapting a frozen multimodal encoder with only lightweight trainable modules, DeCo demonstrates that parameter-efficient learning is also applicable to remote sensing while substantially reducing training costs. This property makes DeCo particularly attractive for high-resolution scenarios, where computational efficiency and visual understanding capability are both critical.

\input{Tables/append_abla}

\section{More Ablation Studies and Qualitative Results}
\label{sec_app_abla}

\textbf{Effects of Task Decoupling and Prior Coupling.} We further investigate the design choices of TSD and HPC in \cref{tab_abla_tsd_hpc}. Removing TSD reduces the average performance to 75.6\%. A straightforward alternative is to duplicate SimFPN for the localization and segmentation branches, which improves the average score to 76.3\%, confirming the benefit of task-specific representations, but at the cost of substantially higher parameters and FLOPs. In comparison, TSD achieves the best average score of 76.6\% with lower computational burden. Removing SWM from TSD decreases the score to 76.2\%, further validating the importance of word conditioned feature decoupling. For prior coupling, removing HPC decreases the average score from 76.6\% to 75.4\%. Introducing only the textual prior improves it to 75.9\%, whereas using only the mask-derived spatial prior yields a larger gain to 76.3\%. In particular, the mask prior improves P@.5:.95 from 69.1\% to 71.1\%, highlighting the value of dense segmentation cues for precise localization. Combining both priors achieves the best overall performance of 76.6\%, demonstrating that sentence-level semantics and mask-derived geometry provide complementary information for cross-task coupling.

\textbf{Effects of the LoRA Scaling Factor $\alpha$.} We then study the effect of the scaling factor $\alpha$ while fixing the rank to $r=32$. As shown in \cref{tab_abla_alpha}, the overall performance gradually improves as $\alpha$ increases from 8 to 32, reaching the best average score of 76.6\% at $\alpha=32$. A smaller $\alpha$ may under-adapt the frozen multimodal encoder, whereas an excessively large value can amplify the low-rank residual and disturb the pretrained representations. This is also reflected by the performance drop to 75.2\% when $\alpha=128$. These results suggest that a moderate adaptation strength is crucial for preserving pretrained knowledge while sufficiently adjusting multimodal representations to the downstream grounding objectives. We therefore set $\alpha=32$ by default.

\begin{figure*}[t]
    \centering
    % \vspace{-5pt}
    \includegraphics[width=0.95\linewidth]{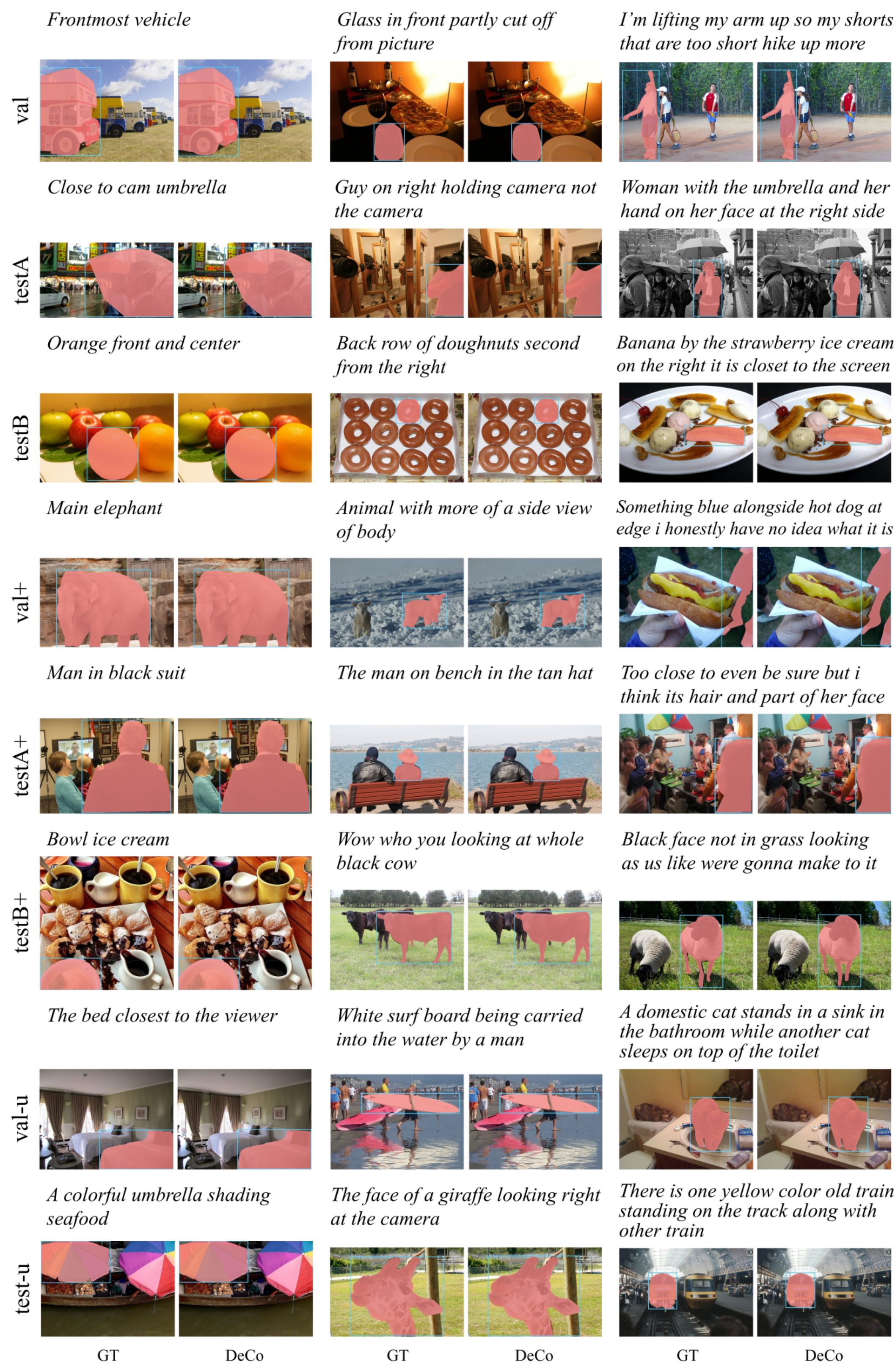}
    \vspace{-5pt}
    \caption{Grounding results of DeCo with BEiT3-L across eight splits under the mixed setting.}
    \label{fig_success_case}
    \vspace{-10pt}
\end{figure*}

\textbf{Grounding across Expression Lengths.} Since longer linguistic expressions typically contain richer attributes and relational cues, we examine the robustness of DeCo to expression length. According to the number of words, the 9,602 samples in G-Ref (test-u) are divided into \textit{short} ($L\leq6$), \textit{middle} ($6<L\leq12$), and \textit{long} ($L>12$) groups, containing 3,415, 4,858, and 1,329 expressions, respectively. As shown in \cref{fig_length_analysis}, we first evaluate the default ablation configuration, DeCo-B under the standard setting (DeCo-B-Standard), which remains stable across different lengths, with no consistent degradation as expressions become longer. Notably, the long group even achieves the highest RES scores of 75.1\% oIoU and 75.6\% mIoU. We further examine our strongest model, DeCo-L under the mixed setting (DeCo-L-Mixed), and observe the same trend. Its REC P@.5 changes only from 93.0\% for short expressions to 92.1\% for long ones, while P@.5:.95 varies from 84.2\% to 83.2\%. Meanwhile, RES oIoU increases from 82.2\% to 83.7\%. These results indicate that DeCo is largely insensitive to expression length and preserves both localization and segmentation quality under increasingly complex descriptions. This robustness is further illustrated by the representative examples in \cref{fig_success_case}. For each length group, we select examples from all eight splits where both REC and RES achieve high IoU. Specifically, we use IoU thresholds of 0.98, 0.95, and 0.90 for short, middle, and long expressions, respectively, to account for the increasing difficulty of processing longer expressions. DeCo accurately identifies the referred targets across different expression lengths, highlighting its consistent performance regardless of expression length.

\begin{figure*}[t]
    \centering
    \vspace{-5pt}
    \includegraphics[width=1.0\linewidth]{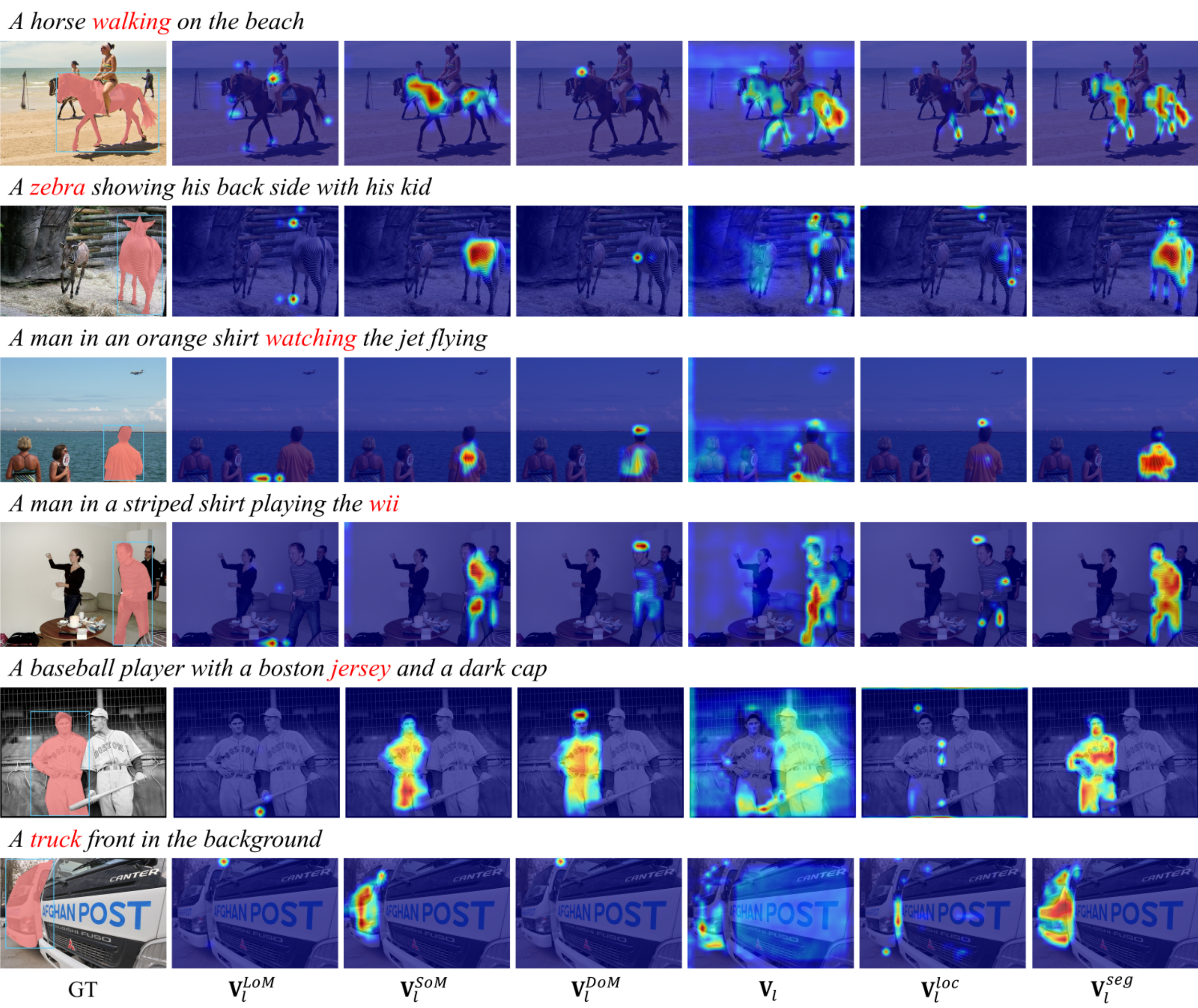}
    \caption{Grad-CAM visualization of task-specific feature preferences and decoupling. LoM ($\mathbf{V}_l^{LoM}$) and SoM ($\mathbf{V}_l^{SoM}$) exhibit distinct localization- and segmentation-oriented responses, while the shared representations of DoM ($\mathbf{V}_l^{DoM}$) and DeCo ($\mathbf{V}_l$) mix these preferences. TSD effectively decouples them into $\mathbf{V}^{loc}_l$ and $\mathbf{V}^{seg}_l$. The salient word selected for modulation is highlighted in red.}
    \label{fig_gradcam}
    \vspace{-10pt}
\end{figure*}

\textbf{Visualization of Task-specific Feature Preferences and Decoupling.} To further confirm the representation conflict between REC and RES, we visualize the third-level ($l=3$) feature responses of different models using Grad-CAM~\citep{gradcam}. As illustrated in \cref{fig_gradcam}, we first compare the task-oriented models and observe different activation patterns: LoM tends to focus on sparse yet discriminative regions that facilitate object localization, whereas SoM distributes its attention more broadly over the target extent to support dense segmentation. This observation provides qualitative evidence that REC and RES favor distinct visual cues. When both objectives share the same representation, as in DoM, the resulting activations tend to either resemble those of one dominant task or exhibit a mixture of the two task preferences. A similar behavior is observed in the shared feature $\mathbf{V}_l$ of DeCo before TSD, demonstrating the limitation of directly coupling different objectives within a common representation. In contrast, after TSD, the decoupled localization-oriented feature $\mathbf{V}^{loc}_l$ recovers the sparse discriminative responses preferred by REC, while the segmentation-oriented feature $\mathbf{V}^{seg}_l$ highlights more complete target regions for RES. Notably, the salient words highlighted in red cover diverse discriminative semantics, including object categories, actions, and descriptive modifiers, rather than being restricted to class attributes. This indicates that the salient-word selection can adaptively identify informative linguistic cues for task-specific feature modulation. These visualizations further support our motivation for decoupling shared representations according to the distinct visual interests of REC and RES before cross-task coupling.

\begin{figure*}[t]
    \centering
    \vspace{-5pt}
    \includegraphics[width=0.95\linewidth]{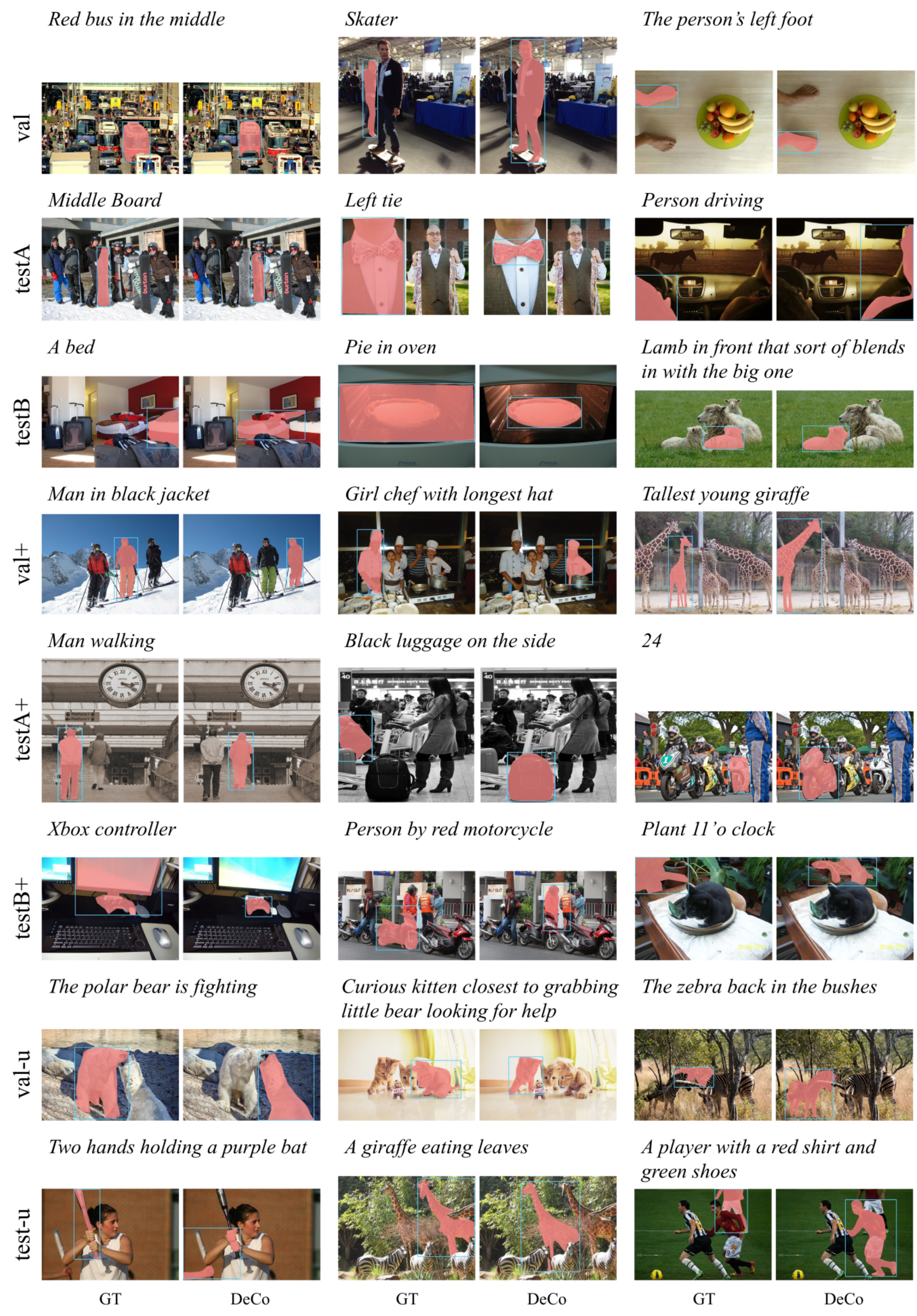}
    \vspace{-5pt}
    \caption{Failure cases of DeCo with BEiT3-L across eight splits under the mixed setting.}
    \label{fig_failure_case}
    \vspace{-10pt}
\end{figure*}

\textbf{Failure Cases.} \cref{fig_failure_case} shows representative failure cases of DeCo with BEiT3-L across eight splits under the mixed setting. We identify a sample as a failure case when its IoU is below 0.3 for both REC and RES. Our analysis reveals three major sources of failure: (i) \textit{Annotation ambiguity} occurs when multiple visual instances can match the same referring expression, making the single-target annotation inherently non-unique; (ii) \textit{Annotation errors} arise when the annotated target does not match the textual description, indicating noise in the ground truth rather than a genuine model failure; (iii) \textit{Model limitations} account for cases where the prediction requires complex logical relations or relies on subtle and fine-grained visual cues. These observations suggest that some apparent failures originate from annotation ambiguity and noise, while the remaining examples highlight the need for stronger relational reasoning and fine-grained visual understanding.

\begin{figure*}[t]
    \centering
    \vspace{-5pt}
    \includegraphics[width=0.92\linewidth]{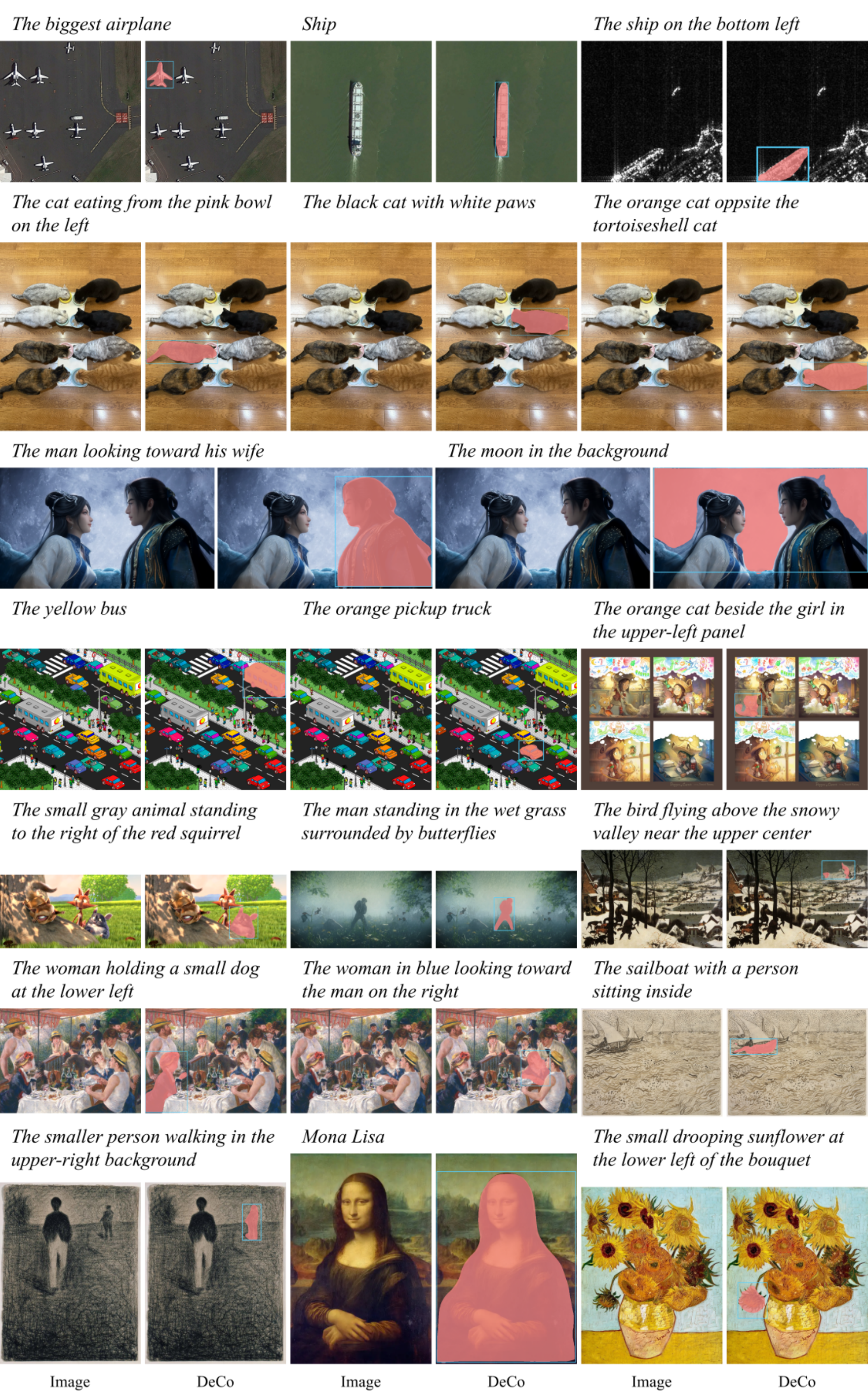}
    \caption{Open-world qualitative results of DeCo with BEiT3-L trained under the mixed setting.}
    \label{fig_open_world}
    \vspace{-10pt}
\end{figure*}

\textbf{Open-World Generalization.} Despite the impressive in-domain performance of DeCo, real-world application requires robust generalization beyond the training distribution. We therefore examine its open-world cognition on unseen images from diverse scenarios, including \textit{optical}, \textit{SAR}, \textit{indoor}, \textit{anime}, \textit{pixel art}, \textit{comics}, \textit{computer graphics}, \textit{painting}, \textit{sketch}, and \textit{artwork}. Free-form referring expressions involving \textit{spatial position}, \textit{visual attributes}, \textit{inter-object relations}, and \textit{logical reasoning} are used to assess fine-grained vision-language alignment under substantial distribution shifts. As shown in \cref{fig_open_world}, DeCo generalizes well across these diverse visual domains. It can distinguish similar instances by fine-grained attributes, localize small targets, and resolve relational descriptions, while remaining effective under strong style shifts from anime and pixel art to sketch and classical artwork. These results indicate that the learned vision-language correspondence transfers beyond the benchmark distributions and retains semantic discrimination and relational reasoning in the wild.

\section{Limitations and Future Works}

Despite its strong performance and generalization, DeCo still has several limitations that motivate further investigation. First, the current formulation assumes single-target referring expressions. The localization branch employs a single object query, while the mask-to-box operation derives only one enclosing box through global spatial extrema, making DeCo less suitable for expressions referring to multiple objects or spatially disconnected regions. A natural extension is to introduce multiple object queries together with instance-aware mask decomposition, where each instance mask is independently converted into a spatial prior. This would enable multiple referred objects to be separately represented, localized, and coupled across REC and RES. Second, although HPC adaptively integrates semantic and mask-derived spatial priors, the quality of the latter still depends on the intermediate segmentation prediction. In challenging cases with severe occlusion, tiny objects, or ambiguous expressions, inaccurate masks may provide imperfect geometric cues. Exploring uncertainty-aware prior estimation or more flexible bidirectional interaction between localization and segmentation could further improve the robustness of cross-task coupling. Finally, parameter-efficient adaptation substantially reduces the number of trainable parameters, but inference still requires forwarding through the frozen multimodal encoder, which may remain computationally demanding for very high-resolution imagery. Future work could investigate token pruning, dynamic-resolution processing, or lightweight multimodal encoders to further reduce inference cost.

%% file: Tables/append_dst_details.tex
\begin{table*}[t]
\centering
% \vspace{-5pt}
\caption{Detailed statistics of the datasets used in our experiments. \textbf{Words} denote the average number of whitespace-separated words in each referring expression. \textbf{REC}, \textbf{RES}, \textbf{RSVG}, and \textbf{RRSIS} represent referring expression comprehension, referring expression segmentation, remote sensing visual grounding, and referring remote sensing image segmentation, respectively.}
\label{tab_dst_details}
\vspace{-0.5em}
\setlength{\tabcolsep}{5pt}
\resizebox{\linewidth}{!}{
\begin{tabular}{l|l|c|c|c|c|c|c|c}
\toprule
\multicolumn{2}{l|}{Datasets}    & Images                 & \multicolumn{4}{c|}{Expressions}                          & Words & Tasks                     \\
\midrule
\multicolumn{9}{l}{\textit{Natural Image Benchmarks}}  \\
\midrule
\multicolumn{2}{l|}{\textbf{RefCOCO}~\citep{refcoco}}     & 19,994                  & train: 120,642 & val: 10,834  & testA: 5,657  & testB: 5,095 & 3.5           & REC, RES                  \\
\midrule
\multicolumn{2}{l|}{\textbf{RefCOCO+}~\citep{refcoco}}    & 19,992                  & train: 120,191 & val: 10,758  & testA: 5,726  & testB: 4,889 & 3.5           & REC, RES                  \\
\midrule
\multirow{2}{*}{\textbf{G-Ref}} & UMD~\citep{gref2}    & \multirow{2}{*}{25,799} & train: 80,512  & val-u: 4,896 & test-u: 9,602 &   /         & 8.4           & \multirow{2}{*}{REC, RES} \\ \cmidrule{4-8}
   & Google~\citep{gref1} &                        & train: 85,474  & val-g: 9,536 &  /           &    /        & 8.4           &                           \\ \midrule
\multicolumn{2}{l|}{\textbf{ReferIt}~\citep{referit}}     & 20,000                  & train: 54,127  & val: 5,842   & test: 60,103  &     /       & 3.4           & REC                       \\
\midrule
\multicolumn{2}{l|}{\textbf{Flickr}~\citep{flickr30k}}      & 31,783                  & train: 427,193 & val: 14,433  & test: 14,481  &    /        & 1.6           & REC                       \\
\midrule
\multicolumn{9}{l}{\textit{Remote Sensing Image Benchmarks}} \\
\midrule
\multicolumn{2}{l|}{\textbf{DIOR-RSVG}~\citep{mgvlf}}   & 17,402                  & train: 26,991  & val: 3,829   & test: 7,500   &    /       & 7.5           & RSVG                      \\
\midrule
\multicolumn{2}{l|}{\textbf{SARVG1.0}~\citep{tacmt}}    & 2,465                   & train: 6,093   & val: 761    & test: 763    &   /         & 6.2           & RSVG                      \\
\midrule
\multicolumn{2}{l|}{\textbf{RRSIS-D}~\citep{rmsin}}     & 17,402                  & train: 12,181  & val: 1,740   & test: 3,481   &  /          & 6.8           & RRSIS                     \\
\midrule
\multicolumn{2}{l|}{\textbf{RIS-LAD}~\citep{ris_lad}}     & 2,103                   & train: 9,712   & val: 1,396   & test: 2,763   &   /        & 19.6          & RRSIS                     \\
\midrule
\multicolumn{2}{l|}{\textbf{RefDIOR}~\citep{rrsecs}}     & 17,380                  & train: 26,824  & val: 3,832   & test: 7,664   &   /        & 7.5           & RSVG, RRSIS               \\
\bottomrule
\end{tabular}}
\end{table*}

%% file: Tables/append_imp_details.tex
\begin{table*}[t]
\centering
% \vspace{-5pt}
\caption{Detailed implementation settings across benchmarks. BS: Batch Size. LR: Learning Rate.}
\label{tab_implem_details}
\vspace{-0.5em}
\setlength{\tabcolsep}{1.5pt}
\resizebox{\textwidth}{!}{
\begin{tabular}{l|l|cccccccccc}
\toprule
\multicolumn{2}{l|}{Settings}
& \textbf{RefCOCO}
& \textbf{RefCOCO+}
& \textbf{G-Ref}
& \textbf{ReferIt}
& \textbf{Flickr}
& \textbf{DIOR-RSVG}
& \textbf{SARVG1.0}
& \textbf{RRSIS-D}
& \textbf{RIS-LAD}
& \textbf{RefDIOR} \\
\midrule

\multicolumn{2}{l|}{Input Size}
& $448{\times}448$
& $448{\times}448$
& $448{\times}448$
& $448{\times}448$
& $448{\times}448$
& $640{\times}640$
& $512{\times}512$
& $512{\times}512$
& $640{\times}640$
& $512{\times}512$ \\

\multicolumn{2}{l|}{Text Length}
& $20$ & $20$ & $20$ & $20$ & $20$
& $20$ & $20$ & $20$ & $40$ & $20$ \\

\midrule

\multirow{2}{*}{BS}
& DeCo-B
& $64$ & $64$ & $64$ & $64$ & $64$
& $32$ & $32$ & $32$ & $32$ & $32$ \\
& DeCo-L
& $32$ & $32$ & $32$ & $32$ & $32$
& $16$ & $16$ & $16$ & $16$ & $16$ \\

\midrule

\multirow{2}{*}{LR}
& DeCo-B
& $2{\times}10^{-4}$
& $2{\times}10^{-4}$
& $2{\times}10^{-4}$
& $2{\times}10^{-4}$
& $2{\times}10^{-4}$
& $1{\times}10^{-4}$
& $1{\times}10^{-4}$
& $1{\times}10^{-4}$
& $1{\times}10^{-4}$
& $1{\times}10^{-4}$ \\

& DeCo-L
& $1{\times}10^{-4}$
& $1{\times}10^{-4}$
& $1{\times}10^{-4}$
& $1{\times}10^{-4}$
& $1{\times}10^{-4}$
& $5{\times}10^{-5}$
& $5{\times}10^{-5}$
& $5{\times}10^{-5}$
& $5{\times}10^{-5}$
& $5{\times}10^{-5}$ \\

\midrule

\multicolumn{2}{l|}{Epochs}
& $35$ & $35$ & $35$ & $35$ & $10$
& $50$ & $90$ & $40$ & $50$ & $50$ \\

\multicolumn{2}{l|}{Milestones}
& $28$ & $28$ & $28$ & $28$ & $8$
& $36, 45$ & $60$ & $30$ & $36, 45$ & $36, 45$ \\

\bottomrule
\end{tabular}}
\end{table*}

%% file: Tables/append_res_miou.tex
\begin{table*}[t]
\centering
% \vspace{-5pt}
\caption{Main results of \textbf{RES} models using \textbf{mIoU} on RefCOCO, RefCOCO+, and G-Ref datasets.}
\label{tab_res_miou}
\vspace{-0.5em}
\setlength{\tabcolsep}{6.5pt}
\resizebox{\linewidth}{!}{
\begin{tabular}{l|c|ccc|ccc|cc|c}
\toprule
\multirow{2}{*}{Models} & Vision\,/\,Language & \multicolumn{3}{c|}{RefCOCO} & \multicolumn{3}{c|}{RefCOCO+} & \multicolumn{2}{c|}{G-Ref} & \multirow{2}{*}{Avg.} \\ \cmidrule{3-10}
 &   Backbone      & val     & testA   & testB   & val+     & testA+    & testB+   & val-u        & test-u        &      \\
\midrule
\multicolumn{11}{l}{\textit{Task-specific Methods (Standard Setting)}}   \\
\midrule

EEVG~\citep{eevg}                    & ViTDet-B\,/\,BERT-B           & 78.2    & 79.3    & \underline{76.6}    & 69.0    & 72.7     & 62.3    & 69.2         & 70.0          & 72.2                 \\
RISCLIP~\citep{risclip}                 & CLIP-B$^*$\,/\,CLIP-B$^*$             & 75.7    & 78.0    & 72.5    & 69.2    & 73.5     & 60.7    & 67.6         & 68.0          & 70.6                 \\
OneRef~\citep{oneref}                  & BEiT3-B\,/\,BEiT3-B                   & 77.6    & 79.1    & 75.1    & 71.3    & 75.4     & 65.5    & 69.4         & 69.7          & 72.9                 \\
MaskRIS~\citep{maskris}                 & Swin-B\,/\,BERT-B             & \underline{78.4}    & \underline{80.2}    & 76.1    & 71.7    & \underline{76.7}     & 64.5    & 69.3         & 69.4          & 73.3                 \\
CMIRNet~\citep{cmirnet}                 & Swin-B\,/\,BERT-B             & 76.0    & 78.1    & 73.1    & 68.2    & 73.0     & 61.3    & 66.7         & 66.5          & 70.4                 \\
DETRIS~\citep{detris}                  & DINOv2-B$^*$\,/\,CLIP-B$^*$           & 76.0    & 78.2    & 73.5    & 68.9    & 74.0     & 61.5    & 67.9         & 68.1          & 71.0                 \\
PropVG~\citep{propvg}                  & BEiT3-B\,/\,BEiT3-B                   & 78.0    & 79.8    & 75.3    & 72.9    & 76.5     & 67.2    & \underline{71.3}         & \underline{72.1}          & \underline{74.1}                 \\
Latent-VG~\citep{latent}               & BEiT3-B\,/\,BEiT3-B                   & 77.7    & 79.5    & 75.5    & \underline{73.2}    & 76.2     & \underline{67.4}    & 71.1         & 71.5          & 74.0                 \\
MiCA~\citep{mica}                    & DINOv2-B$^*$\,/\,CLIP-B$^*$           & 76.6    & 78.9    & 73.8    & 69.8    & 74.7     & 63.5    & 68.7         & 68.8          & 71.9                 \\
AMLRIS~\citep{amlris}                  & Swin-B\,/\,BERT-B             & 77.9    & 79.5    & 75.0    & 71.3    & 75.6     & 64.6    & 69.2         & 69.7          & 72.9                 \\
WIMFRIS~\citep{wimfris}                 & DINOv2-B$^*$\,/\,CLIP-B$^*$           & 77.3    & 79.2    & 74.8    & 70.2    & 75.1     & 63.5    & 69.2         & 69.4          & 72.3                 \\
\rowcolor[HTML]{ECECEC}
\textbf{DeCo (Ours)}              & BEiT3-B$^*$\,/\,BEiT3-B$^*$    & \textbf{80.3} & \textbf{81.8} & \textbf{77.5} & \textbf{75.8} & \textbf{79.2} & \textbf{70.3} & \textbf{74.0} & \textbf{74.5} & \textbf{76.7}   \\

\midrule
RISCLIP~\citep{risclip}                 & CLIP-L$^*$\,/\,CLIP-L$^*$             & 78.9    & 81.5    & 75.4    & 74.4    & 78.8     & 66.8    & 71.8         & 71.7          & 74.9                 \\
OneRef~\citep{oneref}                  & BEiT3-L\,/\,BEiT3-L                   & \underline{80.1}    & \underline{82.2}    & \underline{77.5}    & \underline{75.2}    & \underline{79.4}     & \underline{70.2}    & \underline{73.2}         & \underline{73.8}          & \underline{76.4}                 \\
DETRIS~\citep{detris}                  & DINOv2-L$^*$\,/\,CLIP-B$^*$           & 77.3    & 79.0    & 75.2    & 70.8    & 75.3     & 64.7    & 69.3         & 70.2          & 72.7                 \\
MiCA~\citep{mica}                    & DINOv2-L$^*$\,/\,CLIP-B$^*$           & 78.0    & 79.6    & 75.8    & 72.2    & 76.7     & 65.9    & 70.5         & 70.7          & 73.7                 \\
WIMFRIS~\citep{wimfris}                 & DINOv2-L$^*$\,/\,CLIP-B$^*$           & 78.2    & 79.7    & 76.3    & 71.9    & 76.2     & 67.2    & 70.4         & 71.0          & 73.9                 \\
\rowcolor[HTML]{ECECEC}
\textbf{DeCo (Ours)}  & BEiT3-L$^*$\,/\,BEiT3-L$^*$     & \textbf{83.2} & \textbf{84.4} & \textbf{81.2} & \textbf{80.5} & \textbf{83.1} & \textbf{75.7} & \textbf{78.7} & \textbf{79.0} & \textbf{80.7}  \\

\midrule
\multicolumn{11}{l}{\textit{Task-specific Methods (Mixed Setting)}}   \\
\midrule

PolyFormer~\citep{polyformer}              & Swin-B\,/\,BERT-B             & 76.0    & 77.1    & 73.2    & 70.7    & 74.5     & 64.6    & 69.4         & 69.9          & 71.9                 \\
RISCLIP~\citep{risclip}                 & CLIP-B$^*$\,/\,CLIP-B$^*$             & 76.0    & 78.6    & 71.9    & 69.7    & 74.3     & 61.4    & 69.6         & 69.6          & 71.4                 \\
EEVG~\citep{eevg}                    & ViTDet-B\,/\,BERT-B           & 79.5    & 81.0    & 77.4    & 71.9    & 76.7     & 66.3    & 73.6         & 73.5          & 75.0                 \\
OneRef~\citep{oneref}                  & BEiT3-B\,/\,BEiT3-B                   & 79.8    & 81.9    & 77.0    & 74.7    & 77.9     & 69.6    & 74.1         & 74.9          & 76.2                 \\
CMIRNet~\citep{cmirnet}                 & Swin-B\,/\,BERT-B             & 78.5    & 80.2    & 76.0    & 71.2    & 75.3     & 65.0    & 71.5         & 72.8          & 73.8                 \\
C$^3$VG~\citep{c3vg}                    & BEiT3-B\,/\,BEiT3-B                   & 81.4    & 82.9    & 79.1    & 77.1    & 79.6     & 72.4    & 76.3         & 77.1          & 78.2                 \\
PropVG~\citep{propvg}                  & BEiT3-B\,/\,BEiT3-B                   & 82.0    & 83.6    & 80.0    & 77.1    & 79.8     & 72.2    & 77.0         & 77.7          & 78.7                 \\
Latent-VG~\citep{latent}               & BEiT3-B\,/\,BEiT3-B                   & 81.0    & 82.3    & 79.8    & 76.9    & 79.5     & \underline{73.0}    & 76.1         & 76.5          & 78.1                 \\
SaFiRe~\citep{safire}                  & Swin-B\,/\,BERT-B             & \underline{82.1}    & \underline{83.7}    & \underline{80.1}    & \underline{77.4}    & \underline{80.5}     & 72.9    & \underline{77.3}         & \underline{78.1}          & \underline{79.0}                 \\
InstanceVG~\citep{instancevg}              & BEiT3-B\,/\,BEiT3-B                   & 81.4    & 83.1    & 79.3    & 76.6    & 79.5     & 71.6    & 75.9         & 76.6          & 78.0       \\
\rowcolor[HTML]{ECECEC}
\textbf{DeCo (Ours)}              & BEiT3-B$^*$\,/\,BEiT3-B$^*$   & \textbf{82.8} & \textbf{84.3} & \textbf{80.9} & \textbf{79.3} & \textbf{81.8} & \textbf{74.7} & \textbf{78.2} & \textbf{78.9} & \textbf{80.1}  \\
\midrule
PolyFormer~\citep{polyformer}              & Swin-L\,/\,BERT-B             & 76.9    & 78.5    & 74.8    & 72.2    & 75.7     & 66.7    & 71.2         & 71.2          & 73.4                 \\
RISCLIP~\citep{risclip}                 & CLIP-L$^*$\,/\,CLIP-L$^*$             & 79.5    & 82.1    & 75.8    & 74.9    & 78.9     & 68.1    & 73.5         & 74.5          & 75.9                 \\
UNINEXT~\citep{uninext}                 & ConvNeXt-L\,/\,BERT-B           & 81.2    & 82.4    & 79.1    & 72.6    & 76.3     & 66.8    & 74.1         & 75.4          & 76.0                 \\
OneRef~\citep{oneref}                  & BEiT3-L\,/\,BEiT3-L                   & 81.3    & 83.1    & 79.5    & 76.6    & 80.2     & 73.0    & 75.7         & 76.8          & 78.2                 \\
CMIRNet~\citep{cmirnet}                 & Swin-L\,/\,BERT-B             & 79.0    & 80.7    & 76.6    & 72.5    & 77.0     & 66.9    & 72.6         & 73.2          & 74.8                 \\
DETRIS~\citep{detris}                  & DINOv2-L$^*$\,/\,CLIP-B$^*$           & 81.0    & 81.9    & 79.0    & 75.2    & 78.6     & 70.2    & 74.6         & 75.3          & 77.0                 \\
MiCA~\citep{mica}                    & DINOv2-L$^*$\,/\,CLIP-B$^*$           & 80.8    & 82.7    & 80.1    & 75.6    & 79.3     & 71.5    & 75.2         & 75.7          & 77.6                 \\
InstanceVG~\citep{instancevg}              & BEiT3-L\,/\,BEiT3-L    & \textbf{86.3} & \textbf{87.1} & \textbf{85.3}    & \underline{82.5}    & \underline{84.3}     & \underline{79.2}    & \underline{81.4}         & \underline{82.3}          & \underline{83.5}                 \\
WIMFRIS~\citep{wimfris}                 & DINOv2-L$^*$\,/\,CLIP-B$^*$           & 81.8    & 82.8    & 79.7    & 76.6    & 80.5     & 72.5    & 75.4         & 75.9          & 78.2                 \\
\rowcolor[HTML]{ECECEC}
\textbf{DeCo (Ours)}              & BEiT3-L$^*$\,/\,BEiT3-L$^*$    & \underline{85.6}          & \underline{86.3}          & \underline{84.0}          & \textbf{83.1} & \textbf{85.0} & \textbf{79.7} & \textbf{82.3} & \textbf{82.6} & \textbf{83.6}       \\     
\bottomrule
\end{tabular}}
\end{table*}

%% file: Tables/append_rsvg.tex
\begin{table*}[t]
\centering
% \vspace{-5pt}
\begin{minipage}[t]{1.0\textwidth}
    \centering
    \caption{Main results of \textbf{REC} and \textbf{RSVG} models on the DIOR-RSVG test set.}
    \label{tab_dior_rsvg}
    \vspace{-0.5em}
    \setlength{\tabcolsep}{5pt}
    \resizebox{1.0\columnwidth}{!}{
    \begin{tabular}{l|c|ccccc|cc}
    \toprule
    Models  & Vision / Language & P@.5 & P@.6 & P@.7 & P@.8 & P@.9 & oIoU & mIoU \\
    \midrule
    TransVG~\citep{transvg} & ResNet-50 / BERT-B & 72.4 & 67.4 & 60.1 & 49.1 & 27.8 & 76.3 & 63.6 \\
    VLTVG~\citep{vltvg}   & ResNet-50 / BERT-B & 69.4 & 65.2 & 58.4 & 46.6 & 24.4 & 72.0 & 60.0 \\
    SegVG~\citep{segvg}   & ResNet-50 / BERT-B & 79.6 & 73.7 & 65.0 & 51.1 & 25.7 & 80.9 & 68.5 \\
    SimVG~\citep{simvg}   & BEiT3-B / BEiT3-B & 79.1 & 73.3 & 66.0 & 51.9 & 26.6 & 74.0 & 69.5 \\
    MGVLF~\citep{mgvlf}   & ResNet-50 / BERT-B & 76.8 & 72.7 & 66.7 & 56.4 & 35.1 & 78.4 & 68.0 \\
    LQVG~\citep{lqvg}    & ResNet-50 / BERT-B & 83.4 & 81.0 & 75.9 & 65.5 & 43.5 & 82.2 & 74.0 \\
    FQRNet~\citep{fqrnet}  & ResNet-50 / BERT-B & 77.2 & 73.2 & 67.2 & 56.9 & 35.4 & 78.8 & 68.4 \\
    EG-DINO~\citep{egdino} & ResNet-50 / BERT-B & 83.1 & 80.1 & 75.0 & 63.6 & 42.3 & 81.1 & 73.4 \\
    GLAED~\citep{glaed}   & ResNet-50 / BERT-B & 84.4 & 81.5 & 73.5 & 62.9 & 43.4 & 82.9 & 73.7 \\
    CPVG~\citep{cpvg}    & BEiT3-B / BEiT3-B & 82.6 & 78.4 & 71.9 & 59.1 & 33.6 & 76.9 & 73.2 \\
    CFCVG~\citep{cfcvg}  & Swin-T / BERT-B & 84.7 & 82.2 & 77.6 & 66.3 & 44.7 & 83.3 & 75.2 \\
    HMIT~\citep{hmit}  & HGNetv2 / BERT-B & 84.4 & 82.2 & 78.1 & 68.2 & 45.8 & 83.1 & 75.3 \\
    VisproVG~\citep{visprovg} & Swin-T / BERT-B & 85.2 & \underline{83.0} & \underline{78.2} & \underline{68.2} & \underline{46.9} & \underline{84.7} & 76.0 \\
    DeCo (Ours)  & BEiT3-B$^*$\,/\,BEiT3-B$^*$ & \underline{85.2} & 82.2 & 76.6 & 66.7 & 44.6 & 84.1 & \underline{76.0} \\
    \rowcolor[HTML]{ECECEC}
    \textbf{DeCo (Ours)} & BEiT3-L$^*$\,/\,BEiT3-L$^*$ & \textbf{86.9} & \textbf{84.3} & \textbf{79.0} & \textbf{69.2} & \textbf{47.6} & \textbf{86.2} & \textbf{78.6}  \\
    \bottomrule
    \end{tabular}}
\end{minipage}

\vspace{3.5mm}

\begin{minipage}[t]{1.0\textwidth}
    \centering
    \caption{Main results of \textbf{REC} and \textbf{RSVG} models on the SARVG1.0 test set.}
    \label{tab_sarvg1.0}
    \vspace{-0.5em}
    \setlength{\tabcolsep}{5pt}
    \resizebox{1.0\columnwidth}{!}{
    \begin{tabular}{l|c|ccccc|cc}
    \toprule
    Models  & Vision / Language & P@.5 & P@.6 & P@.7 & P@.8 & P@.9 & oIoU & mIoU \\
    \midrule
    TransVG~\citep{transvg} & ResNet-50 / BERT-B & 82.2 & 76.6 & 69.5 & 50.5 & 15.9 & -    & 68.8 \\
    RefTR~\citep{reftr}   & ResNet-50 / BERT-B & 77.2 & 74.4 & 69.8 & 59.9 & 27.0 & -    & 68.4 \\
    LBYLNet~\citep{lbylnet} & ResNet-50 / BERT-B & 77.1 & 76.4 & 75.9 & 74.8 & 71.6 & -    & 72.7 \\
    VLTVG~\citep{vltvg}   & ResNet-50 / BERT-B & 82.4 & 80.7 & 73.1 & 62.4 & 29.8 & -    & 72.4 \\
    SeqTR~\citep{seqtr}   & ResNet-50 / BERT-B & 85.7 & 84.3 & 80.4 & 69.8 & 34.2 & -    & 74.8 \\
    TACMT~\citep{tacmt}   & ResNet-50 / BERT-B & 89.4 & 88.6 & 87.0 & 83.8 & 68.3 & -    & 82.8 \\
    DeCo (Ours) & BEiT3-B$^*$\,/\,BEiT3-B$^*$  & \underline{89.6} & \underline{89.1} & \underline{88.0} & \underline{84.0} & \underline{73.5} & \underline{80.6} & \underline{83.8} \\
    \rowcolor[HTML]{ECECEC}
    \textbf{DeCo (Ours)} & BEiT3-L$^*$\,/\,BEiT3-L$^*$  & \textbf{91.2} & \textbf{90.2} & \textbf{88.5} & \textbf{85.0} & \textbf{75.1} & \textbf{83.5} & \textbf{86.3}  \\
    \bottomrule
    \end{tabular}}
\end{minipage}

\vspace{3.5mm}

\begin{minipage}[t]{1.0\textwidth}
    \centering
    \caption{Main results of \textbf{REC} and \textbf{RSVG} models on the RefDIOR test set.}
    \label{tab_refdior_rsvg}
    \vspace{-0.5em}
    \setlength{\tabcolsep}{5pt}
    \resizebox{1.0\columnwidth}{!}{
    \begin{tabular}{l|c|ccccc|cc}
    \toprule
    Models  & Vision / Language & P@.5 & P@.6 & P@.7 & P@.8 & P@.9 & oIoU & mIoU \\
    \midrule
    TransVG~\citep{transvg} & Swin-T / BERT-B & 53.7 & 46.3 & 36.1 & 21.3 & 5.3 & 67.1 & 46.7 \\
    VLTVG~\citep{vltvg}  & Swin-T / BERT-B  & 59.4 & 51.5 & 41.2 & 26.4 & 7.8  & 70.2 & 51.7 \\
    PseudoQ~\citep{pseudoq} & Swin-T / BERT-B & 57.1 & 48.7 & 37.6 & 23.1 & 5.4  & 69.2 & 49.7 \\
    QRNet~\citep{qrnet} & Swin-T / BERT-B   & 57.5 & 49.4 & 39.1 & 23.7 & 5.8  & 69.5 & 50.1 \\
    D-MDETR~\citep{dmdetr} & Swin-T / BERT-B & 60.9 & 53.3 & 43.6 & 28.6 & 8.1  & 69.9 & 52.5 \\
    LQVG~\citep{lqvg} & Swin-T / BERT-B    & 78.7 & 75.6 & 70.5 & 60.8 & 42.9 & 81.4 & 70.7 \\
    LPVA~\citep{lpva} & Swin-T / BERT-B    & 60.5 & 52.6 & 41.9 & 27.1 & 7.3  & 71.1 & 52.7 \\
    CCFormer~\citep{rrsecs} & Swin-T / BERT-B & 82.4 & 79.1 & 73.5 & 64.1 & 45.8 & 82.4 & 74.1 \\
    DeCo (Ours) & BEiT3-B$^*$\,/\,BEiT3-B$^*$   & \underline{83.1}   & \underline{79.7}   & \underline{74.2}    & \underline{65.0}    & \underline{47.0}   & \underline{83.1}  & \underline{75.0} \\
    \rowcolor[HTML]{ECECEC}
    \textbf{DeCo (Ours)} & BEiT3-L$^*$\,/\,BEiT3-L$^*$   & \textbf{85.3} & \textbf{82.4} & \textbf{77.3} & \textbf{67.5} & \textbf{50.0} & \textbf{84.2} & \textbf{77.0}  \\
    \bottomrule
    \end{tabular}}
\end{minipage}

\end{table*}

%% file: Tables/append_rrsis.tex
\begin{table*}[t]
\centering
% \vspace{-5pt}
\begin{minipage}[t]{1.0\textwidth}
    \centering
    \caption{Main results of \textbf{RES} and \textbf{RRSIS} models on the RRSIS-D test set.}
    \label{tab_rrsis_d}
    \vspace{-0.5em}
    \setlength{\tabcolsep}{5pt}
    \resizebox{1.0\columnwidth}{!}{
    \begin{tabular}{l|c|ccccc|cc}
    \toprule
    Models  & Vision / Language  & P@.5 & P@.6 & P@.7 & P@.8 & P@.9 & oIoU & mIoU \\
    \midrule
    LAVT~\citep{lavt} & Swin-B / BERT-B  & 69.7 & 63.9 & 54.1 & 42.0 & 25.0 & 77.7 & 61.3 \\
    CrossVLT~\citep{crossvlt} & Swin-B / BERT-B & 66.4 & 59.4 & 49.8 & 38.7 & 23.3 & 75.5 & 58.5 \\
    CARIS~\citep{caris} & Swin-B / BERT-B & 71.5 & 63.5 & 52.9 & 40.9 & 23.9 & 77.2 & 62.2 \\
    DMMI~\citep{dmmi} & Swin-B / BERT-B & 68.7 & 61.0 & 50.3 & 38.4 & 21.6 & 76.2 & 60.1 \\
    LGCE~\citep{lgce} & Swin-B / BERT-B  & 69.9 & 64.0 & 53.6 & 40.8 & 24.2 & 77.2 & 60.9 \\
    RMSIN~\citep{rmsin} & Swin-B / BERT-B  & 74.2 & 67.1 & 55.6 & 41.7 & 23.1 & 77.5 & 64.0 \\
    FIANet~\citep{fianet} & Swin-B / BERT-B & 74.5 & 67.0 & 56.3 & 42.8 & 24.1 & 76.9 & 64.0 \\
    LSCF~\citep{lscf} & Swin-B / BERT-B & 74.3 & 67.7 & 56.3 & 43.1 & 25.7 & 77.4 & 64.3 \\
    CroBIM-U~\citep{crobim} & ConvNeXt-B / BERT-B & 75.6 & 67.7 & 56.5 & 42.6 & 24.2 & 76.7 & 65.1 \\
    VSPNet~\citep{vspnet} &  MambaVision / CLIP & 75.1 & 67.6 & 55.4 & \underline{44.6} & 25.4 & 77.8 & 64.7 \\
    S2CLNet~\citep{s2lcnet} & Swin-B / BERT-B & 74.9 & 68.9 & 56.7 & 44.1 & 25.6 & 77.7 & 64.5 \\
    MCD-Net~\citep{mcdnet} & Swin-B / BERT-B & 75.6 & 68.7 & 57.5 & 44.2 & \textbf{26.0} & 78.1 & 65.1 \\
    DeCo-B (Ours) & BEiT3-B$^*$\,/\,BEiT3-B$^*$  & \underline{77.3} & \underline{70.4} & \underline{58.0} & 43.5 & 23.9 & \underline{78.3} & \underline{66.4} \\
    \rowcolor[HTML]{ECECEC}
    \textbf{DeCo-L (Ours)} & BEiT3-L$^*$\,/\,BEiT3-L$^*$  & \textbf{81.3} & \textbf{74.6} & \textbf{61.3} & \textbf{45.8} & \underline{25.7} & \textbf{79.7} & \textbf{68.9}  \\
    \bottomrule
    \end{tabular}}
\end{minipage}

\vspace{3.5mm}

\begin{minipage}[t]{1.0\textwidth}
    \centering
    \caption{Main results of \textbf{RES} and \textbf{RRSIS} models on the RIS-LAD test set.}
    \label{tab_ris_lad}
    \vspace{-0.5em}
    \setlength{\tabcolsep}{5pt}
    \resizebox{1.0\columnwidth}{!}{
    \begin{tabular}{l|c|ccccc|cc}
    \toprule
    Models  & Vision / Language  & P@.5 & P@.6 & P@.7 & P@.8 & P@.9 & oIoU & mIoU \\
    \midrule
    LAVT~\citep{lavt} & Swin-B / BERT-B   & 31.7 & 27.0 & 23.0 & 17.6 & 8.6  & 42.0 & 30.1 \\
    ASDA~\citep{asda} & Swin-B / BERT-B  & 33.2 & 26.9 & 20.2 & 12.8 & 3.9  & 37.5 & 33.3 \\
    LGCE~\citep{lgce} & Swin-B / BERT-B  & 26.4 & 21.8 & 17.6 & 13.3 & 7.1  & 40.8 & 26.2 \\
    RMSIN~\citep{rmsin} & Swin-B / BERT-B & 43.4 & 38.1 & 32.4 & 25.1 & 13.8 & 48.8 & 39.6 \\
    FIANet~\citep{fianet} & Swin-B / BERT-B  & 40.2 & 35.2 & 29.9 & 23.7 & 13.2 & 43.4 & 37.4 \\
    CADFormer~\citep{cadformer} & Swin-B / BERT-B & 42.2 & 37.0 & 31.9 & 23.6 & 12.6 & 46.5 & 39.3 \\
    SAARN~\citep{ris_lad} & Swin-B / BERT-B & 45.0 & 39.3 & 33.4 & 26.2 & 15.3 & 49.6 & 41.7 \\
    DeCo (Ours)  & BEiT3-B$^*$\,/\,BEiT3-B$^*$ & \underline{51.1}          & \underline{46.3}          & \underline{40.8}          & \underline{31.9}          & \underline{15.5}          & \underline{50.3}          & \underline{47.1}          \\
    \rowcolor[HTML]{ECECEC}
    \textbf{DeCo (Ours)} & BEiT3-L$^*$\,/\,BEiT3-L$^*$ & \textbf{56.1} & \textbf{51.6} & \textbf{45.9} & \textbf{35.5} & \textbf{18.9} & \textbf{54.8} & \textbf{50.2}   \\
    \bottomrule
    \end{tabular}}
\end{minipage}

\vspace{3.5mm}

\begin{minipage}[t]{1.0\textwidth}
    \centering
    \caption{Main results of \textbf{RES} and \textbf{RRSIS} models on the RefDIOR test set.}
    \label{tab_refdior_rrsis}
    \vspace{-0.5em}
    \setlength{\tabcolsep}{5pt}
    \resizebox{1.0\columnwidth}{!}{
    \begin{tabular}{l|c|ccccc|cc}
    \toprule
    Models & Vision / Language  & P@.5 & P@.6 & P@.7 & P@.8 & P@.9 & oIoU & mIoU \\
    \midrule
    LAVT~\citep{lavt}  & Swin-T / BERT-B   & 67.8 & 62.1 & 54.2 & 44.7 & 29.9 & 79.0 & 60.8 \\
    CGFormer~\citep{cgformer} & Swin-T / BERT-B & 67.8 & 61.8 & 54.9 & 45.3 & 30.1 & 79.3 & 61.3 \\
    DMMI~\citep{dmmi}  & Swin-T / BERT-B   & 70.5 & 64.6 & 57.0 & 47.0 & 31.0 & 79.9 & 63.3 \\
    ReMamber~\citep{remamber} & Swin-T / BERT-B & 71.4 & 66.1 & 58.9 & 48.7 & 32.9 & 80.2 & 63.9 \\
    MagNet~\citep{magnet} & Swin-T / BERT-B  & 72.5 & 66.7 & 58.8 & 49.0 & 33.1 & 80.8 & 64.8 \\
    LGCE~\citep{lgce}  & Swin-T / BERT-B   & 69.8 & 64.0 & 56.6 & 46.9 & 31.2 & 79.6 & 62.3 \\
    RMSIN~\citep{rmsin}  & Swin-T / BERT-B  & 68.6 & 62.9 & 55.4 & 45.8 & 30.5 & 79.0 & 61.6 \\
    CCFormer~\citep{rrsecs} & Swin-T / BERT-B & 80.1 & 74.9 & 66.6 & 54.8 & 36.8 & 80.9 & 71.0 \\
    DeCo-B (Ours)  & BEiT3-B$^*$\,/\,BEiT3-B$^*$  & \underline{81.6} & \underline{76.0} & \underline{67.2} & \underline{55.3} & \underline{36.8} & \underline{81.8} & \underline{71.6} \\
    \rowcolor[HTML]{ECECEC}
    \textbf{DeCo-L (Ours)} & BEiT3-L$^*$\,/\,BEiT3-L$^*$  & \textbf{83.2} & \textbf{77.4} & \textbf{69.5} & \textbf{57.8} & \textbf{38.6} & \textbf{83.8} & \textbf{73.4}  \\
    \bottomrule
    \end{tabular}}
\end{minipage}

\end{table*}

%% file: Tables/append_abla.tex
\begin{figure*}[t]
\centering
% \vspace{-10pt}

\begin{minipage}[t]{0.56\textwidth}
\centering
    \footnotesize
    \captionof{table}{Ablation on different feature decoupling and prior coupling. SWM: Salient Word-level Modulation.}
    \label{tab_abla_tsd_hpc}
    \vspace{-0.5em}
    \resizebox{1.0\columnwidth}{!}{
    \begin{tabular}{l|rr|cc|cc|c}
    \toprule
    \multirow{2}{*}{\diagbox[width=9.3em,height=2.8em]{Models}{\shortstack{test-u}}} 
    & \multicolumn{2}{c|}{Trainable Head} & \multicolumn{2}{c|}{REC} & \multicolumn{2}{c|}{RES} & \multirow{2}{*}{Avg.} \\ \cmidrule{2-7}
    & Params.  & FLOPs  & P@.5  & P@.5:.95  & oIoU   & mIoU  &   \\ 
    \midrule
    DeCo w/o TSD & 9.3M & 17.6G & 86.5 & 70.5 & 72.2 & 73.3 & 75.6 \\
    w/ Dual SimFPN & 12.1M & 28.1G & 86.8 & \underline{71.3} & 72.8 & 74.2 & 76.3 \\
    w/ TSD (w/o SWM) & 10.4M & 22.0G & \underline{86.8} & 71.2 & 72.6 & 74.2 & 76.2 \\
    \midrule
    \rowcolor[HTML]{ECECEC}
    \textbf{DeCo} & 11.6M & 24.2G & \textbf{87.0} & \textbf{71.7} & \textbf{73.0} & \textbf{74.5} & \textbf{76.6} \\
    \midrule
    DeCo w/o HPC & 10.6M & 24.2G & 86.3 & 69.1 & 72.5 & 73.6 & 75.4 \\
    w/ HPC (Text) & 11.6M & 24.2G & 86.5 & 70.2 & 72.7 & 74.2 & 75.9 \\
    w/ HPC (Mask) & 10.6M & 24.2G & 86.8 & 71.1 & \underline{72.8} & \underline{74.4} & \underline{76.3} \\
    \bottomrule
    \end{tabular}
    }
\end{minipage}
\hspace{0.5mm}
\begin{minipage}[t]{0.42\textwidth}
\centering
    \footnotesize
    \captionof{table}{Ablation on the LoRA scaling factor $\alpha$.}
    \label{tab_abla_alpha}
    \vspace{-0.5em}
    \resizebox{1.0\columnwidth}{!}{
        \begin{tabular}{l|cc|cc|c}
    \toprule
    \multirow{2}{*}{\diagbox[width=5em,height=2.7em]{$\alpha$}{\shortstack{test-u}}} 
    & \multicolumn{2}{c|}{REC}
    & \multicolumn{2}{c|}{RES}
    & \multirow{2}{*}{Avg.} \\
    \cmidrule{2-5}
    & P@.5 & P@.5:.95 & oIoU & mIoU & \\
    \midrule
    \multicolumn{6}{c}{$r=32$, Trainable Backbone Param.: 4.7M, FLOPs: 1.9G}  \\
    \midrule
    8   & \underline{86.7} & 70.8 & 72.6 & 73.9 & 76.0 \\
    16  & 86.5 & 71.4 & \textbf{73.1} & \underline{74.2} & \underline{76.3} \\
    \rowcolor[HTML]{ECECEC}
    \textbf{32} & \textbf{87.0} & \textbf{71.7} & \underline{73.0} & \textbf{74.5} & \textbf{76.6} \\
    64  & 86.3 & \underline{71.4} & 72.6 & 74.0 & 76.1 \\
    128 & 85.5 & 70.2 & 71.8 & 73.3 & 75.2 \\
    \bottomrule
    \end{tabular}
    }	
\end{minipage}

\vspace{2mm}

\includegraphics[width=0.95\linewidth]{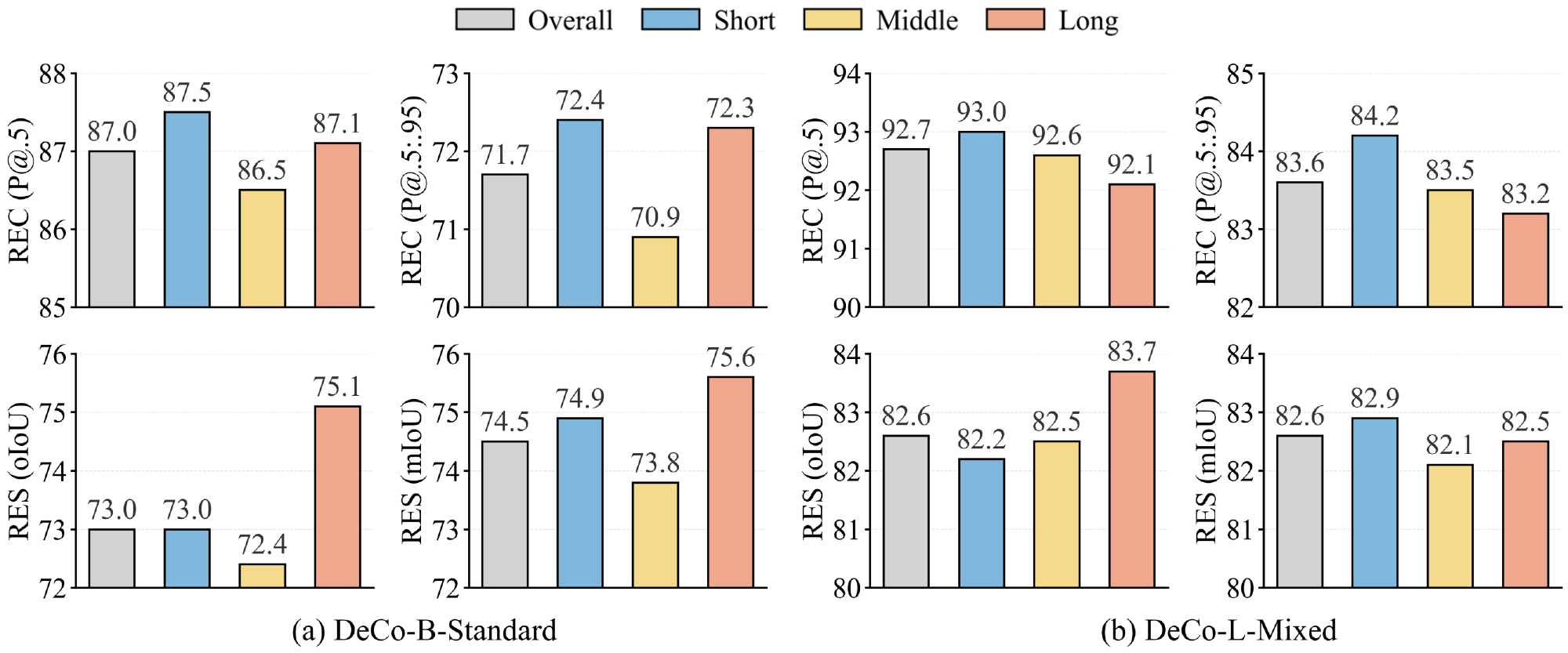}
% \vspace{-5pt}
\captionof{figure}{Performance across different expression lengths on G-Ref (test-u).}
\label{fig_length_analysis}
% \vspace{-10pt}
\end{figure*}